%% file: main.tex
\documentclass[10pt,twocolumn,letterpaper]{article}
\usepackage{authblk}
\usepackage{makecell}
\usepackage{svg}
\usepackage[T1]{fontenc}
\usepackage[pagenumbers]{cvpr} 

\usepackage{colortbl}  
\usepackage{booktabs}  
\definecolor{cvprblue}{rgb}{0.21,0.49,0.74}
\definecolor{text_green}{rgb}{0.43,0.68,0.3}
\definecolor{text_yellow}{rgb}{0.99,0.76,0.18}
\usepackage[pagebackref,breaklinks,colorlinks,allcolors=cvprblue]{hyperref}

\def\paperID{14360} 
\def\confName{CVPR}
\def\confYear{2025}

\title{Text-to-Edit: Controllable End-to-End Video Ad Creation via Multimodal LLMs}

\author[]{Dabing Cheng, Haosen Zhan, Xingchen Zhao, Guisheng Liu, Zemin Li, \\Jinghui Xie, Zhao Song, Weiguo Feng, Bingyue Peng}
\affil[]{ByteDance Inc}

\affil[]{{\tt\small \textbf{\url{https://text2edit.github.io}}}}

\begin{document}
\maketitle
\input{preamble}


\vspace{-1.2em}
\input{sec/0-abstract}    
\input{sec/1-introduction}

\input{sec/2-relatedwork}
\input{sec/3-methodolody}
\input{sec/4-experiment}
\input{sec/5-conclusion}

{
    \small
    \bibliographystyle{ieeenat_fullname}
    \bibliography{main}
}

\input{sec/X_suppl}

\end{document}

%% file: preamble.tex
%
%
\newcommand{\red}[1]{{\color{red}#1}}
\newcommand{\todo}[1]{{\color{red}#1}}
\newcommand{\TODO}[1]{\textbf{\color{red}[TODO: #1]}}
\newcommand{\Zhao}[1]{{[\color{red}Zhao:#1]}}
\newcommand\fast{\mathrm{fast}}
\newcommand\slow{\mathrm{slow}}
\newcommand\clip{\mathrm{clip}}

%% file: sec/0-abstract.tex
\begin{abstract}

    The exponential growth of short-video content has ignited a surge in the necessity for efficient, automated solutions to video editing, with challenges arising from the need to understand videos and tailor the editing according to user requirements. Addressing this need, we propose an innovative end-to-end foundational framework, ultimately actualizing precise control over the final video content editing. Leveraging the flexibility and generalizability of Multimodal Large Language Models (MLLMs), we defined clear input-output mappings for efficient video creation. To bolster the model's capability in processing and comprehending video content, we introduce a strategic combination of a denser frame rate and a slow-fast processing technique, significantly enhancing the extraction and understanding of both temporal and spatial video information. Furthermore, we introduce a text-to-edit mechanism that allows users to achieve desired video outcomes through textual input, thereby enhancing the quality and controllability of the edited videos. Through comprehensive experimentation, our method has not only showcased significant effectiveness within advertising datasets, but also yields universally applicable conclusions on public datasets.
\end{abstract}

%% file: sec/1-introduction.tex
\begin{figure*}[t]
    \centering
    \includegraphics[width=0.95\textwidth]{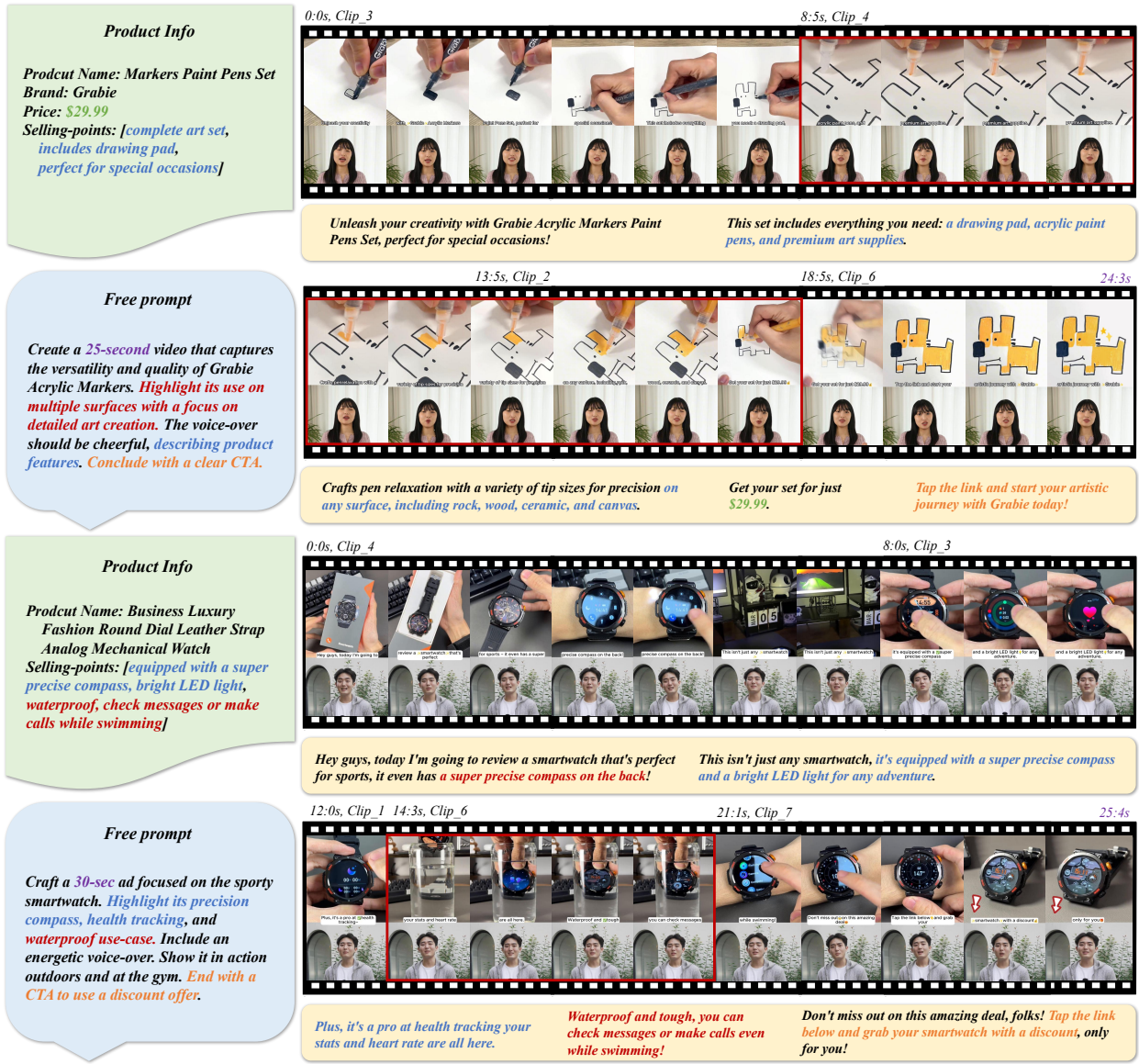} 
    \caption{Examples of videos generated by our method. Leveraging the Multimodal LLM, we proposed an end-to-end solution for creating advertising narrative videos. This approach directly outputs a draft protocol for video edits by processing inputs including product information, expected requirements (free prompt), and video clips. Importantly, it facilitates precise and controllable video editing that perfectly aligns with user-defined free prompts. For details, refer to the figure where corresponding colors indicate aligned information. See more cases in Appendix \ref{app-cases}.}
    \label{fig:first_figure}
    \vspace{-1.2em}
\end{figure*}

\vspace{-1.2em}
\section{Introduction}
With the rapid rise of short videos online, their role in advertising and marketing has grown significantly, emphasizing the need for smarter, more efficient methods to accelerate high-quality video ad production. Recent advancements in Multimodal Large Language Models (MLLMs)~\cite{alayrac2022flamingo,zhu2023minigpt,liu2024visual,zhang2023internlm} show promising potential in commercial advertising, particularly in generating marketing scripts \cite{yang2024synchronized} and understanding ad images or videos, which substantially improves production efficiency and reduces costs in creating video ads.

However, current MLLM applications in advertising video creation are limited to specific production stages, lacking an end-to-end framework.
We systematically analyze the underlying challenges and summarize them into the following three points:
(1) Video creation involves complex processes and material management, including tasks like video clips arrangement, script generation, background music recommendation, and their synchronization with timeline alignment. This demands not only a clear and universally applicable definition of inputs and outputs but also requires significant model flexibility and collaboration.
(2) Editing video materials requires a deep understanding of content, focusing on both spatial information and critical temporal dynamics, which are vital for the logic of the video storyline. However, existing models~\cite{maaz2023video, luo2023valley, zhang2023video, lin2023video} for video understanding generally capture overall content but miss crucial temporal variations.
(3) The content generated in video production must maintain both diversity and controllability to ensure richness and practical relevance. Imposing rules on model outputs limits diversity, while increasing diversity often produces outputs that fail to meet practical needs, reducing production efficiency.

To address these challenges, we propose a unified video creation solution based on multimodal LLMs for narrative advertising videos. 
Leveraging the flexibility and generalizability of LLMs, we clearly defined the input and output. 
Specifically, we input product information, textual user editing requirements, and clips of video materials. 
The multimodal model understands the input information and completes the video creation task, directly generating a JSON-formatted editing draft detailing clips arrangement track, voice-over script track, and decorative-related tags in the video. 
Finally, we perform simple post-processing and rendering on the draft to complete the video production.

To tackle the video understanding problems, we propose to employ a higher frame density by sampling video frames at up to 2 frames per second (fps), harnessing the self-attention mechanism to enhance sensitivity to temporal variations. To facilitate the input of longer videos, we reduce the number of tokens per frame to prevent token overflow. Following token compression, our system can support the input of up to 600 frames of video, a capability rarely seen in other works. Additionally, to avoid losing spatial semantic signals, we balance the need for detailed spatial information and temporal dynamics through a slow-fast strategy~\cite{feichtenhofer2019slowfast,huang2024lita}. 
In this approach, the slow pathway processes frames sparsely (e.g., 0.5fps) with more visual tokens per frame (e.g., 16 or 64 tokens) to capture detailed spatial information. 
Meanwhile, the fast pathway processes frames densely (e.g., 2fps) with fewer tokens (e.g., 1 or 4 tokens), focusing on temporal dynamics to maintain consistency.

Building on LLMs’ inherent diversity to enhance video quality and control, we introduce a text-to-edit approach, allowing users to specify desired outputs through text input. We set boundaries for managing free-prompting and use a data enhancement pipeline to simulate user needs. As shown in Fig.~\ref{fig:first_figure}, our framework, trained on high-quality pairwise free-prompt data, closely aligns with user inputs, ensuring high output quality and extensive control over video editing.

Overall, our core contributions can be summarized as follows:

\begin{itemize}
    \item We propose a novel video editing framework with a multimodal LLM that clearly defines input-output formats, streamlining production by managing material comprehension and arrangement in one step. To the best of our knowledge, this is the first end-to-end MLLM-based video editing framework.
    \item We introduce an effective approach employing a denser frame rate and a slow-fast processing strategy, which significantly enhances the model's ability to extract and understand temporal and spatial video information.
    \item To improve the controllability of the model's output, we develop a text-driven video editing method that ensures the final output aligns precisely with user expectations.
    \item We conduct comprehensive experiments and exploratory analyses, demonstrating that our proposed video editing framework can optimally perform advertising short-video editing tasks. Moreover, our framework generalizes well to public datasets, achieving applicable results.
\end{itemize}

%% file: sec/2-relatedwork.tex
\begin{figure*}[t]
    \centering
    \includegraphics[width=0.95\textwidth]{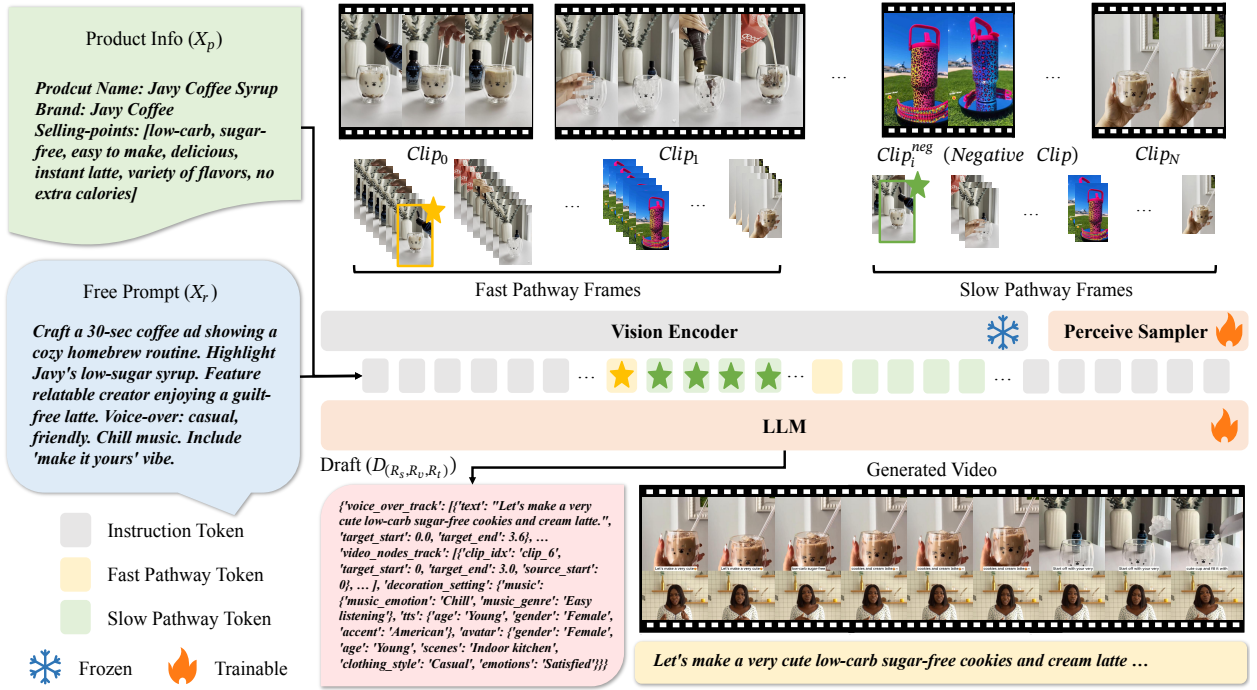} 
    \caption{Illustration of our model architecture. We input product information, free prompt, and video clips into the model, generating a JSON-formatted video editing draft that is transformed into a video through post-processing and rendering. We implement a slow-fast dual-pathway strategy for frame rates: the fast pathway has a higher frame rate with fewer tokens per frame (\textcolor{text_yellow}{\textbf{yellow star}}), and the slower pathway has a lower frame rate with more tokens per frame (\textcolor{text_green}{\textbf{green star}}). By integrating the free prompt into instructions, we enhance the flexibility and control of the video editing process.}
    \label{fig:model_architecture}
    \vspace{-0.8em}
\end{figure*}

\vspace{-0.2em}
\section{Related Work}

\subsection{Multimodal Large Language Models}
Recently, a growing focus on MLLMs such as Flamingo~\cite{alayrac2022flamingo}, MiniGPT-4~\cite{zhu2023minigpt}, LLaVA~\cite{liu2024visual}, LLaVA-NeXT series~\cite{liu2024llavanext, zhang2024llavanextvideo, li2024llavanext-interleave}, InternLM-XComposer series~\cite{zhang2023internlm, internlmxcomposer2}, InternVL~\cite{chen2024internvl} and MiniCPMV~\cite{yao2024minicpm} has expanded their potenial.
Initial studies mainly centered on the alignment of image and text modalities, with a strong emphasis on image understanding. 
More recent work has shifted to video understanding~\cite{maaz2023video, luo2023valley, zhang2023video, lin2023video, wang2024internvideo2, maaz2024videogpt+, zhang2024llavanextvideo, huang2024lita}. 
Early studies, however, were limited by their reliance on basic sparse frame (usually, 8 or 16 frames are extracted from a single video) extraction strategies~\cite{maaz2023video, luo2023valley, zhang2023video, lin2023video} and simplistic temporal feature aggregation modules~\cite{wang2024internvideo2, maaz2024videogpt+}, which proved inadequate for fine-grained video analysis. 
Recognizing this limitation, recent studies have demonstrated that utilizing more video frames \cite{zhang2024llavanextvideo} as input or more effective temporal information memory module~\cite{song2024moviechat} yields substantial improvements. 
In our video editing task, it is essential to incorporate denser frames and input images with dual frame rates, including both slow and fast~\cite{feichtenhofer2019slowfast, huang2024lita}, to provide the model with a comprehensive and nuanced representation of the content.

\subsection{Video Editing and Composition}
Automated video editing and composition has consistently attracted interests from researchers \cite{chua1995video, ahanger1998automatic}. 
However, many existing works faces two main limitations. First, they often focus on addressing challenges within specific sub-processes of editing, such as classifying and organizing materials, retrieving materials based on the script, or supplementing text based on materials. For instance, works like Quickcut \cite{truong2016quickcut}, B-script \cite{huber2019b}, Write-A-Video \cite{wang2019write}, and Transcript-to-Video \cite{xiong2022transcript} rely heavily on pre-written or re-wrtiten scripts to organize or retrieve materials required for editing. 
Other frameworks, specifically, Argaw \textit{et al.}~\cite{argaw2022anatomy} solely facilitate video shots sorting, while Yang \textit{et al.}~\cite{yang2024synchronized} focus on generating advertising oral script for given shots, and ChunkyEdit \cite{leake2024chunkyedit} is designed to assist editors in organizing video interview clips.
Second, other works are limited by their specific application contexts. For example, Arev\textit{et al.}~\cite{arev2014automatic} primarily address editing within multi-camera social settings, while Leake \textit{et al.}~\cite{leake2017computational} focus on dialogue-driven scenarios.
In contrast, while our focus is primarily on short narrative advertising videos, our proposed method is highly adaptable to a broad range of video editing scenarios. 

%% file: sec/3-methodolody.tex
\vspace{-0.2em}
\section{Methodolody}

\subsection{Definition of Task}

Our model input comprises three primary aspects: product information, user-specified free prompt and segmented video clips.
More specifically, the product information, represented by $X_{p}$, includes the product name, brand, price, and key selling points.
For free prompts, the model adheres to user-specified directives such as video storyline, script patterns or decorative elements, encapsulated by $X_{r}$.
More importantly, the input video clips are demonstrated as $C=\{ \clip_{0}, \cdots, \clip_{i}, \cdots, \clip_{N} \}$. 
Here, $N$ signifies the quantity of the clips, and $i$ denotes the clip's index. 
Moreover, each clip segment can be represented by the extracted frames: $\clip_{i}= \{ \mathrm{fr}_0,\cdots, \mathrm{fr}_l, \cdots, \mathrm{fr}_{L}  \}$, where $\mathrm{fr}_l$ denotes a specific frame within the clip, and $L$ represents the total number of frames in the clip.
To closely mimic real scenarios, we incorporated interference clips (denoted by $\{ \clip_{i}^{ \mathrm{neg} }, \cdots \} $) from other videos as negative examples. 
This strategy aims to discourage the model from selecting irrelevant clips during editing, thereby enhancing its ability to resist interference.
Based on empirical values, we employ a rounded Gaussian distribution ($\mathrm{num}\mathrm{Negative}\mathrm{Clip} = \max \{ 0, \mathrm{round} (\mathcal{N}(2.5, 8) ) \}$) to sample interference clips.

Upon processing through our framework, the model directly outputs a simplified JSON draft of the video protocol, represented as $D_{(R_s,R_v,R_t)}$, primarily consisting of three tracks: voice-over track $R_{s}$, video nodes track $R_{v}$, and decoration setting $R_{t}$. We define our task as follow: 
\begin{align}\label{eq:task_difine} 
    D_{(R_s,R_v,R_t)} = \mathcal{M}(X_p, X_r, C),
\end{align}
where $\mathcal{M}(\cdot)$ is our proposed model. 
After basic post-processing and converting the generated JSON into an actual video draft, then it can be rendered into a video.
Refer to the Appendix \ref{app-draft} for details on the post-processing and conversion of the draft.

\subsection{Denser Slow-fast Strategy}
As depicted in the Fig.~\ref{fig:model_architecture}, the process begins by obtaining video clips. 
To increase the model's sensitivity to long-term changes in videos, we employ denser frame sampling frequency (with a maximum rate of 2 fps) and reduce the number of tokens per (to a minimum of 1 token) frame to prevent excessive token overflow.
More specifically, We first sample each video $\clip_i$ to extract denser frames $\tilde{C}_{i}$ as follow:
\begin{align}\label{eq:sample_difine} 
\tilde{C}_{i} =
\begin{cases}
    \mathrm{fr}_{L//2} ,& \text{ if } t_i < \frac{1}{f}, \\
    \mathrm{Uniform} (  \clip_i, \frac{Lf}{t_i} ), & \text{ otherwise},
\end{cases}
\end{align}
where $\mathrm{Uniform}(\cdot)$ indicates that frames are uniformly sampled at intervals of $\frac{L \cdot f}{t_i}$, and $t_i$ is the duration of $\clip_{i}$, $L$ denotes the total number of frames in $\clip_{i}$, $f$ represents the expected sampling fps, when $t_i < \frac{1}{f}$, we take the middle frame of the clip frames. 

Subsequently, each set of acquired clip frames is processed by a vision encoder $\mathrm{Enc}(\cdot)$ and perceive sampler $\mathrm{PerS}(\cdot)$, transforming the visual frames into a sequence of visual tokens.
The perceive sampler can generally be divided into two types: Query based~\cite{zhang2023internlm, yao2024minicpm} (e.g. Q-former~\cite{li2023blip}) and Multi-Layer Perceptron (MLP)~\cite{chen2024internvl, liu2024llavanext, zhang2024llavanextvideo, li2024llavanext-interleave} based. 
Accordingly, for these two different perceive sampler methods, we adopt two distinct strategies for compressing visual tokens. 
For a Q-former based approach, we control the token count for each frame by squeezing the number of query tokens. 
For a MLP based approach, we compress the image feature map size after the visual encoder using an average 2D pooling method. 
This video encoding process can be expressed with the following 
formula:
\begin{align}\label{venc equation}
    \mathrm{PerS}(\tilde{C}_{i})=
    \begin{cases}
        \mathrm{Query}( \mathrm{Squeeze} (q' ), \mathrm{Enc}( \tilde{C}_{i}) ),& \\
        \mathrm{MLP}( \mathrm{Pooling}( \mathrm{Enc} (\tilde{C}_{i}))),&
    \end{cases}
\end{align}
where $q' \in {\mathbb{R}^{s' \times d}}$ and $s'=s/\alpha$, $s$ is the number of query tokens initialized by the pre-trained model, $\alpha$ is squeeze coefficient.
During training, we reinitialize $s'$ query tokens and reload the the pre-trained weights which been squeezed by average pooling with an interval of $\alpha$.

Moreover, to simultaneously accommodate long temporal information and spatial semantic signals, we adopt a slow-fast pathway strategy. 
The encoder operates on two distinct fps pathways and can be rewritten as:
\begin{align}\label{slow_fast_enc} 
    E^{\fast} = \mathrm{PerS} (\tilde{C}^{\fast}), \\
    E^{\slow} = \mathrm{PerS} (\tilde{C}^{\slow}),
\end{align}
where $E^{\fast}$ and $E^{\slow}$ represent different pathways visual tokens respectively.

\textbf{Slow Pathway Frames:} This pathway operates at a lower frame rate where each frame occupies more tokens, effectively capturing abundant spatial semantic information. 
In the context of the slow pathway, where we designate the frame rate as $f$ and the duration of a certain $\clip_i$ as $t_i$, each frame is encode into $k$ tokens. 
Subsequently, we are able to derive that this particular segment yields $e_i^{\slow}$ tokens:
\begin{align}\label{eq:slow_equation}
    e_i^{\slow}=f\cdot k  \cdot t_i .
\end{align}

\textbf{Fast Pathway Frames:} In contrast, this pathway features a higher frame rate, encoding visual data where each frame occupies fewer tokens, providing a more detailed representation of rapidly changing scenes. 
Assume that the $\clip_i$ in the fast pathway will be a high frame rate $\beta \cdot f$, 
with the same duration $t_i$, each frame will be encoded as $k/\beta$ tokens, and finally we can get $e_i^{\fast}$ tokens:
\begin{align}\label{eq:fast_equation}
    e_i^{\fast} = \beta f \cdot k  \cdot t_i /\beta = f \cdot k \cdot t_i .
\end{align}

Thus, from Eq.~\eqref{eq:slow_equation} and Eq.~\eqref{eq:fast_equation}, it is evident that the number of tokens encoded per segment correlates directly with the video length. 
Consequently, this method adeptly handles videos of varying durations, seamlessly integrating extensive information on temporal changes and ample spatial semantics.

Finally, these visual tokens, combined with text tokens derived from product information $X_p$ and specific requirements $X_r$, are encoded by $\mathrm{Emb}(\cdot)$ and then input into the LLM, which subsequently produces the output video draft.
\begin{align}
    D_{(R_s,R_v,R_t)} = \mathrm{LLM}(E^{\fast}, E^{\slow}, \mathrm{Emb}(X_p), \mathrm{Emb}(X_r)).
\end{align}

\subsection{Text-Driven Controllable Editing}
In practical applications, users have specific objectives for advertising videos, such as emphasizing key selling points or adhering to particular editing routines. When models fail to meet these requirements, content may become formulaic and fall short of expectations, diminishing effectiveness. To enhance model output controllability, we propose the use of free-prompt, enabling users to specify their requirements for tailored content. Defining the scope of free-prompt capabilities is essential, therefore, we identify key dimensions, including video duration, visual storyline, target audience, script routine, emphasis on selling points, avatar imagery, Text-to-Speech (TTS) timbre, and music style.

\begin{figure}[t]
    \centering
    \includegraphics[width=0.9\columnwidth]{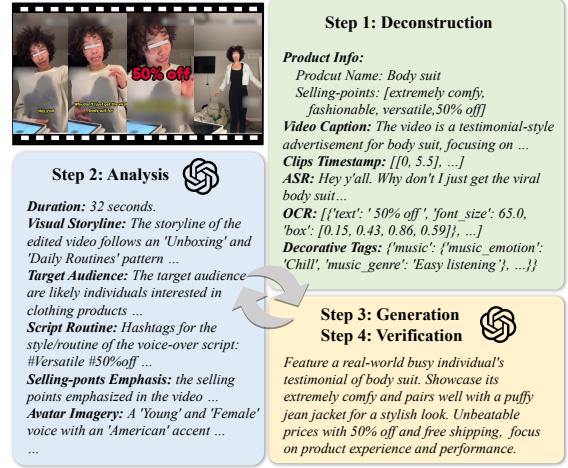} 
    \caption{Pipeline for generating free-prompt. We produce high-quality free-prompt by the four-step process: deconstruction, analysis, generation, and verification. All experiments in this paper utilize the "gpt-4o-2024-08-06" version.}
    \label{fig:freeprompt}
    \vspace{-0.8em}
\end{figure}

As shown in Fig.~\ref{fig:freeprompt}, we follow four key steps to construct high-quality free-prompt training data.
\textbf{Step 1: Deconstruction}, we begin with a comprehensive deconstruction of the original video. Initially, Automatic Speech Recognition (ASR) and Optical Character Recognition (OCR) are used to extract the voice-over script and subtitle stickers. Then, we utilize TransNet \cite{soucek2024transnet} to determine the timestamps of the video clips. Sparse frames are extracted to generate video captions using MiniCPMV \cite{yao2024minicpm}. Leveraging the collected data, we utilize GPT-4o \cite{achiam2023gpt} to recommend appropriate decorative element labels (see Appendix \ref{app-tags} for details) for the original video.
\textbf{Step 2: Analysis}, utilizing the deconstructed data along with predefined support dimensions, we employ GPT-4o to analyze and determine specific content for each dimension. For example, as shown in the Fig.~\ref{fig:freeprompt}, it is necessary to identify that the visual storyline pertains to a product user experience.
\textbf{Step 3: Generation}, following the analysis, a free-prompt is generated to simulate real user requirements. To improve the generalizability, we randomly omit some dimensions during the requirement generation process.
\textbf{Step 4: Verification}, since the generated free-prompt may contain inaccuracies, we implement a verification and revision mechanism. This involves reviewing and correcting the free-prompt, resuling in a 50.13\% improvement in the free-prompt following (FPF) rate.

\label{free_prompt_construct}

%% file: sec/4-experiment.tex
\section{Experiments}

\subsection{Data Collection}

We compiled a dataset of 100K short videos centered around product marketing, predominantly featuring daily small commodities such as beauty products, food and beverages, and electronics, named as VideoAds dataset.
Out of these, we designated 2K videos as the testing set and utilized the remaining for training. 
When constructing the data sample, we synthesize the instruction as shown in Fig.~\ref{fig:model_architecture}. Additionally, for the expected draft output, we assemble the video draft by generating the ASR, video clips, and decorative tags deconstructed in Section \ref{free_prompt_construct}. 
Specifically, the ASR corresponds to the voice-over track of the draft. 
However, the text recognized by ASR sometimes contains errors and lacks punctuation, so we use GPT-4o for correction, subsequently restoring it to the voice-over track with timestamp for each sentence. 
For video clips, we incorporate negative sample clips to shuffle the order and record the correct clip sorting index to construct the video nodes track. 
Regarding the decorative element labels, we use the deconstructed labels as the content for the decorative settings.
For detailed sample construction templates, see the Appendix \ref{app-data}.

Additionally, we utilized the Shot2story \cite{han2023shot2story20k} dataset to evaluate our method's shot sorting ability and shot caption generation capability on a public dataset, aiming to verify that our approach can be seamlessly applied to other video editing scenarios with minimal modifications.

\begin{table*}[t]
    \centering
    \resizebox{0.95\linewidth}{!}{
        \begin{tabular}{lccccccccc}
            \toprule
            \textbf{Model} & \textbf{Visual Encoder} & \textbf{Setting(fps/token)} & \textbf{CRA$ \uparrow $}  & \textbf{CSA$ \uparrow $} & \textbf{FPF$ \uparrow $} & \textbf{VSR$ \uparrow $} & \textbf{SQ$ \uparrow $} & \textbf{DTPR$ \uparrow $} \\
            \midrule
            GPT-4o & - & 0.125/- & 10.80 & 38.03 & 84.21 & 23.18 & 76.93 & 66.48/75.89 \\
            \midrule
            \multicolumn{9}{l}{\textit {\textbf{Denser Frame}}} \\
            \midrule
            LLaVA-NeXT-Vicuna-7B \cite{chiang2023vicuna} & CLIP \cite{radford2021learning}+MLP & 0.125/1 & 28.37 & 55.64 & - & - & - & - \\
            LLaVA-NeXT-Vicuna-7B & CLIP+MLP & 2/1 & 29.24 & 55.36 & 89.74 & 23.23 & 74.44 & 78.92/79.72  \\
            LLaVA-NeXT-Vicuna-7B & CLIP+MLP  & 0.125/4 & 29.13 & 55.68 & - & - & - & -  \\
            LLaVA-NeXT-Vicuna-7B & CLIP+MLP & 2/4 & 32.07 & 76.25 & 90.07 & 23.53 & 74.21 & \textbf{80.54/80.28}  \\
            InternLM-XComposer-7B \cite{zhang2023internlm} & EVA-CLIP \cite{sun2023eva}+Q-former & 0.125/1 & 42.19 & 89.65 & - & - & - & - \\
            InternLM-XComposer-7B & EVA-CLIP+Q-former & 2/1 & \textbf{44.45} & \textbf{89.76} & 90.88 & 23.51 & 76.17 & 78.26/79.34  \\
            InternLM-XComposer-7B & EVA-CLIP+Q-former & 0.125/4 & 39.49 & 87.24 & - & - & - & -  \\
            InternLM-XComposer-7B & EVA-CLIP+Q-former & 2/4 & 40.72 & 80.02 & 89.90 & 23.42 & 75.48 & 77.88/78.53  \\
            LLaVA-NeXT-Mistral-7B \cite{jiang2023mistral} & CLIP+MLP & 2/1 & 31.08 & 61.71 & 90.14 & 23.33 & 76.75 & 78.25/79.20  \\
            LLaVA-NeXT-Mistral-7B & CLIP+MLP & 2/4 & 43.96 & 88.43 & \textbf{91.00} & 23.59 & 77.35 & 78.92/79.72  \\
            LLaVA-NeXT-Qwen-7B \cite{bai2023qwen} & SigLIP \cite{zhai2023sigmoid}+MLP & 2/1 & 29.43 & 58.90 & 89.97 & 23.22 & \textbf{78.29} & 77.69/79.70  \\
            LLaVA-NeXT-Qwen-7B & SigLIP+MLP & 2/9 & 39.66 & 85.84 & 90.72 & 23.54 & 78.11 & 78.12/80.22  \\
            InternVL-26B \cite{chen2024internvl} & InternViT \cite{chen2024internvl}+MLP & 2/1 & 41.69 & 85.22 & 90.47 & 23.61 & 75.85 & 79.68/80.14  \\
            InternVL-26B & InternViT+MLP & 2/4 & 43.58 & 88.22 & 90.86 & \textbf{23.62} & 76.33 & 79.61/79.92  \\
            MiniCPMV2.6-7B \cite{yao2024minicpm} & SigLIP+Query & 2/1 & 44.20 & 79.59 & 86.42 & 22.87 & 72.20 & 78.17/77.58  \\
            MiniCPMV2.6-7B & SigLIP+Query & 2/4 & 38.73 & 73.01 & 86.77 & 23.13 & 70.14 & 77.40/77.02  \\
            \midrule
            \multicolumn{9}{l}{\textit {\textbf{Slow-fast Strategy}}} \\
            \midrule
            LLaVA-NeXT-Vicuna-7B & CLIP+MLP & \textit {fast:}2/4 & 32.07 & 76.25 & 90.07 & 23.53 & 74.21 & 80.54/80.28  \\
            LLaVA-NeXT-Vicuna-7B & CLIP+MLP & \textit {slow:}0.5/16 & 35.33 & 81.97 & 89.39 & 23.53 & 73.86 & 80.52/80.30  \\
            LLaVA-NeXT-Vicuna-7B & CLIP+MLP & \textit {slow:}0.125/64 & 37.50 & 83.14 & 89.46 & 23.61 & 73.57 & 80.52/80.27  \\
            \rowcolor[gray]{0.93}  
            LLaVA-NeXT-Vicuna-7B & CLIP+MLP & \textit {fast:}2/4 \textit {slow:}0.5/16 & 37.01 & 82.37 & \textbf{90.37} & 23.61 & \textbf{74.96} & \textbf{80.74/80.72}  \\
            \rowcolor[gray]{0.93}  
            LLaVA-NeXT-Vicuna-7B & CLIP+MLP & \textit {fast:}2/4 \textit {slow:}0.125/64 & \textbf{43.94} & \textbf{86.93} & 89.63 & \textbf{23.97} & 72.97 & 80.33/80.26  \\
            \midrule
            InternLM-XComposer-7B & EVA-CLIP+Q-former & \textit {fast:}2/4 & 40.72 & 80.02 & 89.90 & 23.42 & 75.48 & 77.88/78.53  \\
            InternLM-XComposer-7B & EVA-CLIP+Q-former & \textit {slow:}0.5/16 & 38.76 & 78.89 & 89.55 & 23.41 & 74.11 & 78.04/78.82  \\
            InternLM-XComposer-7B & EVA-CLIP+Q-former & \textit {slow:}0.125/64 & 34.61 & 70.44 & 89.44 & 23.37 & 73.77 & 77.87/78.21  \\
            \rowcolor[gray]{0.93}  
            InternLM-XComposer-7B & EVA-CLIP+Q-former & \textit {fast:}2/4 \textit {slow:}0.5/16 & 43.91 & 87.77 & 90.32 & \textbf{23.56} & 74.87 & \textbf{78.26/79.27}  \\
            \rowcolor[gray]{0.93}  
            InternLM-XComposer-7B & EVA-CLIP+Q-former & \textit {fast:}2/4 \textit {slow:}0.125/64 & \textbf{45.91} & \textbf{89.53} & \textbf{90.41} & 23.55 & \textbf{74.94} & 77.91/79.02  \\
            \bottomrule
        \end{tabular}
    }
    \caption{Performance analysis of different models using denser frame, and slow-fast strategy. We report the fps and the number of tokens encoded per frame, with the best performance marked in bold.}
    \label{tab:results}
    \vspace{-0.8em}
\end{table*}

\begin{figure}[h]
    \centering
    \includegraphics[width=0.95\columnwidth]{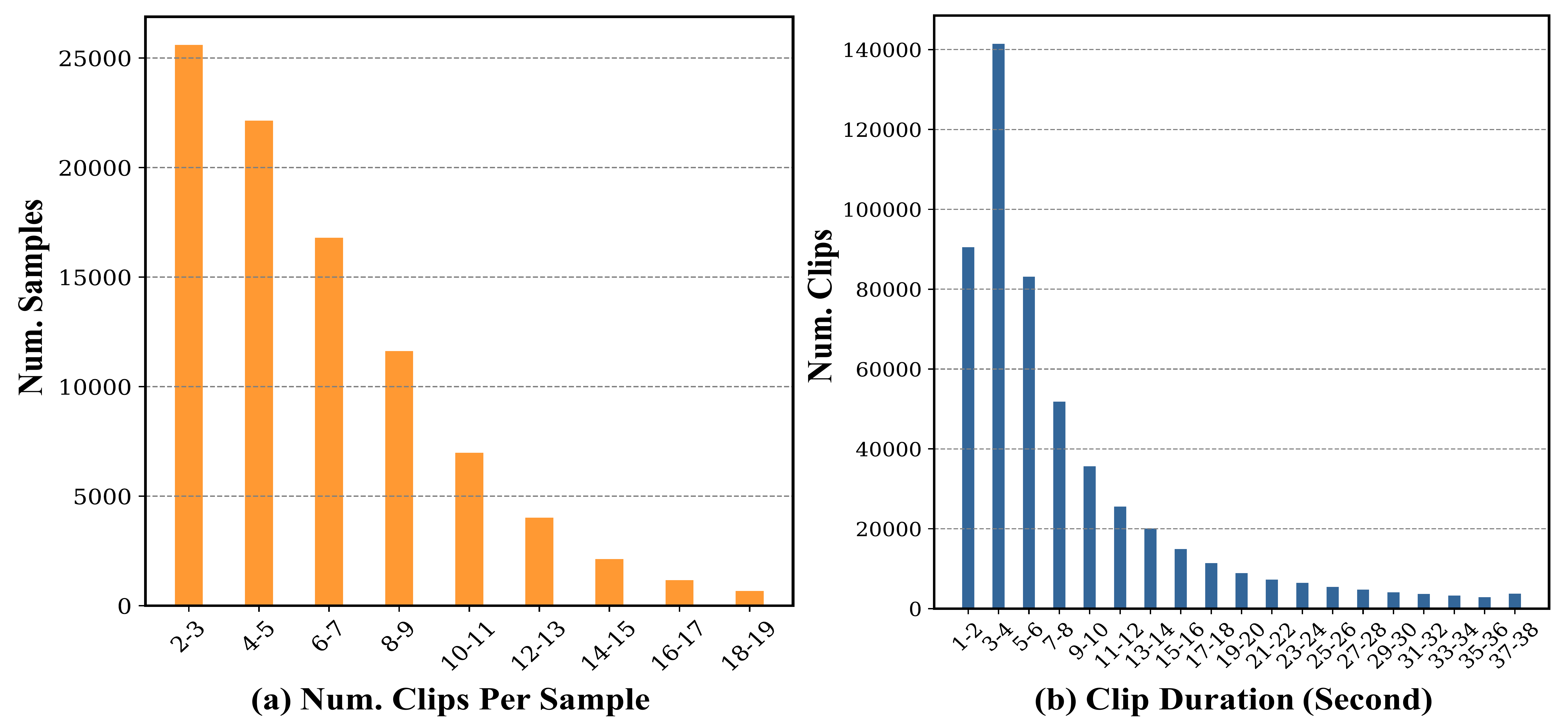} 
    \caption{Depiction of our VideoAds dataset distribution.~\textbf{(a)} illustrates the distribution of the number of clips per data sample, encompassing both positive and negative clips, with a mean $\mu_{\mathrm{numClips}}=5.88$.~\textbf{(b)} depicts the distribution of clip durations, with a mean $\mu_{\mathrm{clipDuration}}=8.03$ seconds.}
    \label{fig:dataset}
    \vspace{-0.8em}
\end{figure}

\subsection{Metrics Definition}
\label{metrics}
Since our proposed task is explicitly defined within the video editing domain, employing the LLM metrics (such as BLEU~\cite{papineni2002bleu}, ROUGE-L~\cite{lin2004rouge}, METEOR~\cite{banerjee2005meteor} and CIDEr~\cite{vedantam2015cider}) commonly used in NLP does not offer significant value (but still report result in the Appendix \ref{app-llm}). 
Furthermore, drawing from our extensive practical experience, we define the evaluation indicators for this task as follows:

\textbf{Clips Rank Accuracy (CRA):}
Measuring the accuracy of predicted video clip sequences, and it can be written as $\mathrm{CRA} = \frac{N_{\mathrm{correct}}}{N_{\mathrm{total}}} \times 100\%$,
where $N_{\text{correct}}$ is the number of samples with all clips in the correct order, and $N_{\mathrm{total}}$ is the total number of samples.

\textbf{Clips Selection Accuracy (CSA):} Assessing the model's accuracy in selecting positive clips, this can be written as 
$\mathrm{CSA} = \frac{N_{\mathrm{select}}}{N_{\mathrm{total}}} \times 100\%$,
and where $N_{\mathrm{select}}$ is the number of samples that do not contain any negative clips and $N_{\mathrm{total}}$ is the total number of samples.

\textbf{Free Prompt Following (FPF):} Evaluating the model's ability to adhere to a free prompt using GPT-4o evaluation and assessed dimensions as section~\ref{free_prompt_construct}.

\textbf{Visual Script Relevance (VSR):} Assessing the correlation between visual imagery and the spoken script using the EMScore \cite{shi2022emscore}, a higher score signifies stronger relevance.

\textbf{Script Quality (SQ):} Examining script quality using GPT-4o to evaluate four dimensions: Baisc, Native Language \& Tone, Touch the Audience, and Creative Narrative.

\textbf{Decorative Tags Precision / Recall (DTPR):} Measures the precision and recall of recommended decorative element tags, including Avatar, TTS Timbre, and Music.
So these can be defined as: $\mathrm{Precision} = \frac{N_{\mathrm{TP}}}{N_{\mathrm{TP}} + N_{\mathrm{FP}}} \times 100\%$, 
$\mathrm{Recall} = \frac{N_{\mathrm{TP}}}{N_{\mathrm{TP}} + N_{\mathrm{FN}}} \times 100\%$,
where $N_{\mathrm{TP}}$, $N_{\mathrm{FP}}$ and $N_{\mathrm{FN}}$ are the number of tags for correctly, incorrectly and should been recommended but is not, respectively.

See more details for GPT-4o evalutation prompts in Appendix~\ref{app-prompts}.
In the Shot2story dataset, we use CRA and CSA to evaluate shots ranking and selection, and NLP metrics for caption generation.

\subsection{Training Details}
In this work, we utilized several various multimodal models, including InternLM-XComposer~\cite{zhang2023internlm}, InternVL \cite{chen2024internvl}, LLaVA-NeXT series~\cite{liu2024llavanext, zhang2024llavanextvideo, li2024llavanext-interleave}, and MiniCPMV \cite{yao2024minicpm}, 
coupled with vision encoders such as CLIP \cite{radford2021learning}, EVA-CLIP~\cite{sun2023eva}, SigLIP \cite{zhai2023sigmoid}, and InternViT-6B~\cite{chen2024internvl}. Base language models included InternLM-7B \cite{team2023internlm}, InternLM2-20B~\cite{cai2024internlm2}, 
and models from the LLaVA-NeXT series such as Vicuna-7B \cite{chiang2023vicuna}, Mistral-7B \cite{jiang2023mistral}, along with Qwen variants~\cite{bai2023qwen, yang2024qwen2}.
Experiments utilized 8 NVIDIA A100 GPUs across 40K iterations with the AdamW optimizer \cite{loshchilov2017decoupled} and a cosine learning rate decay \cite{loshchilov2016sgdr}, starting at a rate of $2 \times e^{-5}$. 
During Supervised Fine-Tuning (SFT), the vision encoder weights were frozen. Meanwhile, the perceive sampler and the weights of the LLMs were actively fine-tuned to optimize performance.

\subsection{Automatic Evaluation Results}
As shown in Table \ref{tab:results}, we use the metrics described in Section~\ref{metrics} to evaluate various models and configurations. 
Initially, we use GPT-4o as baseline model (detailed in the Appendix~\ref{app-gpt4pipe}). 
Due to the lack of fine-tuning, it exhibits significant disadvantages compared to fine-tuned models. For example, under the same 0.125 fps configuration (64 token setting for LLaVA-NeXT-Vicuna-7B, later simplified as Vicuna), GPT-4o underperforms Vicuna in most metrics, particularly in CRA (-26.70) and CSA (-45.11). 
However, benefiting from its strong general text generation ability, GPT-4o performs better than Vicuna in SQ (+3.36). 
Furthermore, we provide results for more multimodal models, which highlight that the InternLM-XComposer model achieves the highest performance in terms of the CRA (44.45) and CSA (89.76) metrics, 
and Mistral, InternVL, Qwen, and Vicuna perform best in FPF (91.00), VSR (23.62), SQ (78.29), and DTPR (80.54/80.28), respectively.

\begin{figure*}[t]
    \centering
    \begin{minipage}[b]{0.34\linewidth}
        \centering
        \includegraphics[width=0.98\linewidth]{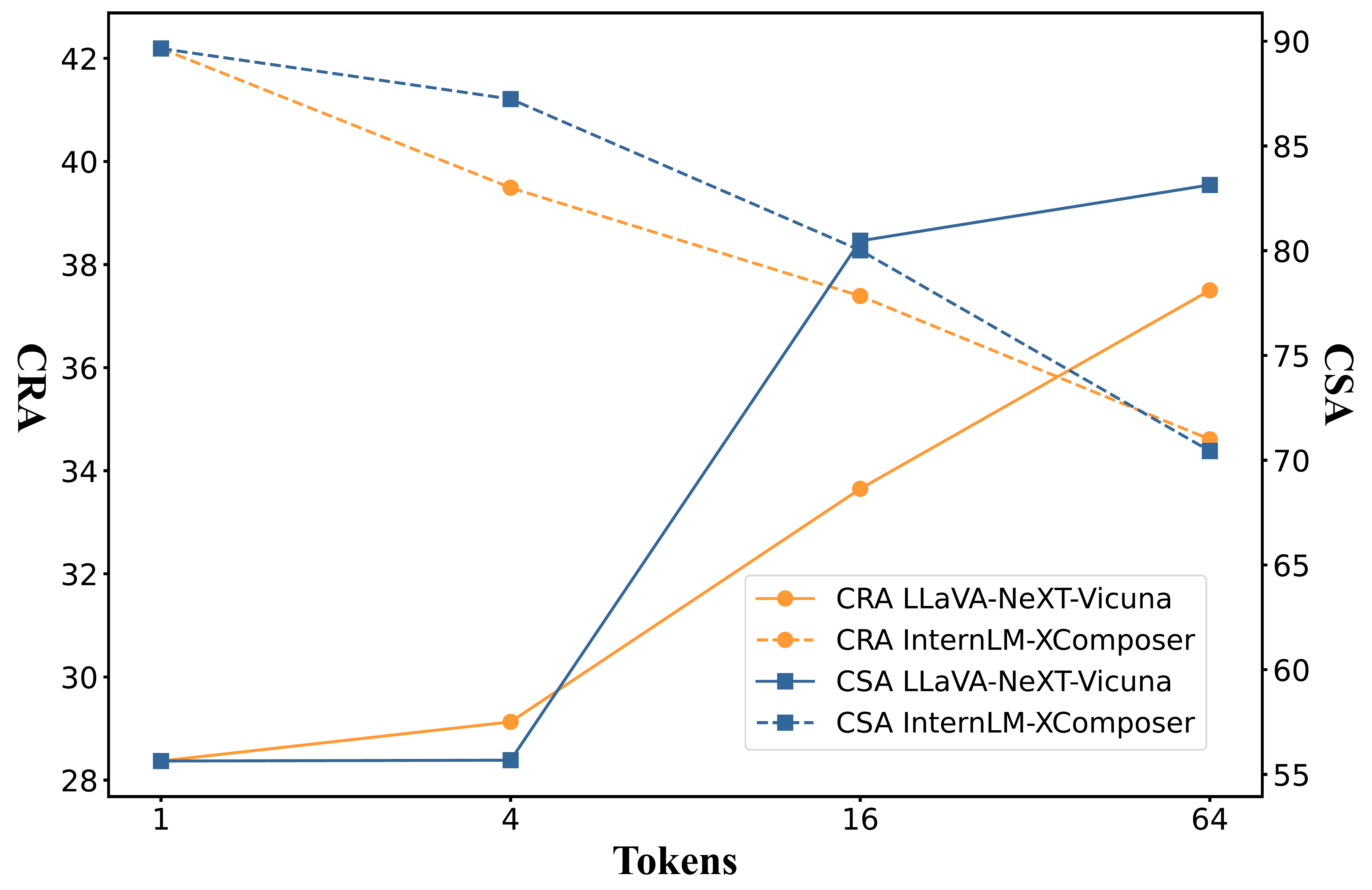}
        \caption{\centering Analyzing the impact of token \newline numbers on CRA and CSA metrics \newline (fps=0.125).}
        \label{fig:effect_token_size}
    \end{minipage}%
    \hfill
    \begin{minipage}[b]{0.33\textwidth}
        \centering
        \includegraphics[width=\textwidth]{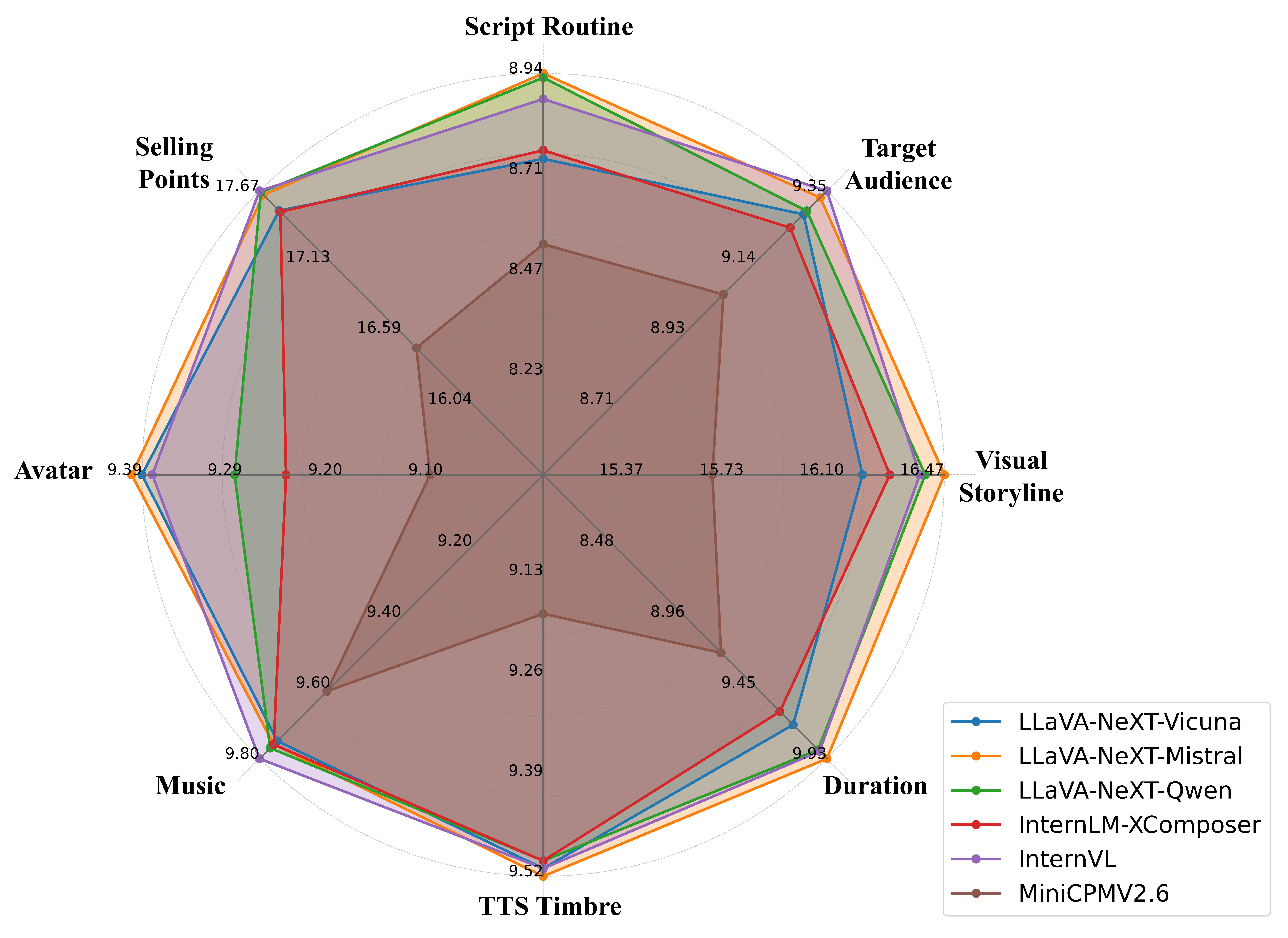}
        \caption{\centering Evaluating different models \newline performance on free-prompt following \newline (Qwen, fps/token:2/9, others, fps/token:2/4).}
        \label{fig:free_prompt_following}
    \end{minipage}%
    \hfill
    \begin{minipage}[b]{0.32\linewidth}
        \centering
        \includegraphics[width=0.98\linewidth]{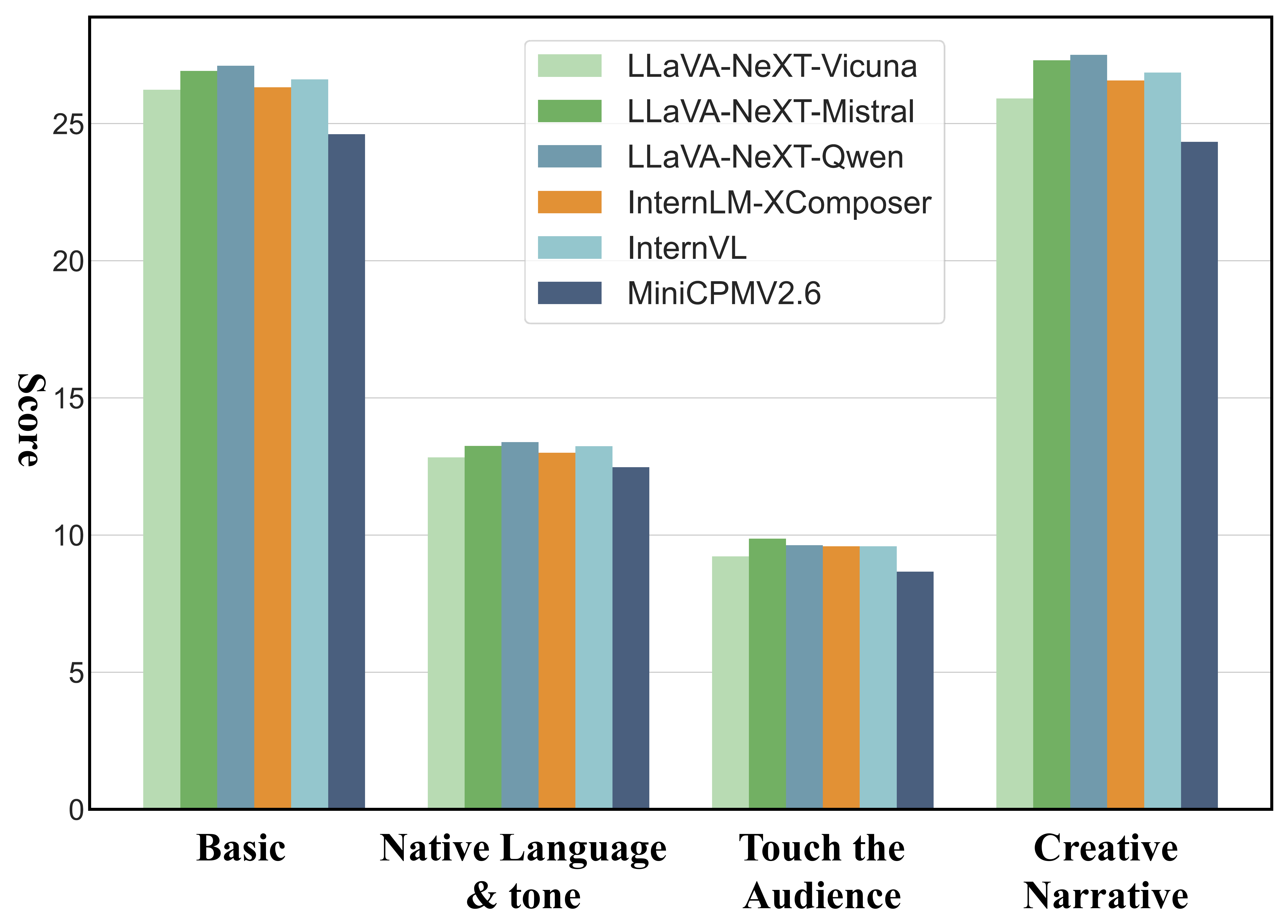}
        \caption{\centering Evaluating different models \newline performance on script quality (Qwen, \newline fps/token:2/9, others, fps/token:2/4).}
        \label{fig:script_quality}
    \end{minipage}
    \label{fig:comparison_metrics}
    \vspace{-0.8em}
\end{figure*}

\textbf{Verification of the Denser Slow-Fast Strategy.}
In the context of the Denser Frame strategy, from the results of the Vicuna and InternLM-XComposer, it is evident that the performance (CRA) at 2 fps surpasses that at 0.125 fps, regardless of whether the token setting is 1 or 4. 
This observation suggests that increasing the fps with a denser frame sampling effectively improves the model's temporal understanding of video sequences. 
More importantly, in the context of the Slow-fast Strategy, we also experimentally verified the effectiveness of the slow-fast strategy. Both Vicuna and InternLM-XComposer, with two sets of slow-fast experiment settings (fast: 2/4 + slow: 0.5/16 and fast: 2/4 + slow: 0.125/64), use the slow-fast strategy consistently outperforms using only slow or fast pathways separately.
For example, for the InternLM-XComposer model, while the fast: 2/4 setting alone reached CRA 40.72 and CSA 80.02, integrating the slow pathway (slow: 0.125/64) elevated performance to CRA 45.91 and CSA 89.53, with improvements of approximately 5.19 and 9.51 points, respectively.
This indicates that balancing detailed spatial information from the slow pathway and temporal dynamics from the fast pathway is crucial for optimal performance. 
From Fig.~\ref{fig:effect_token_size}, we observe that increasing token numbers per frame from 1 to 64 results in opposing trends for CRA and CSA in InternLM-XComposer (dropping from 42.19 and 89.65 to 34.61 and 70.44, respectively) and LLaVA-NeXT-Vicuna (increasing from 28.37 and 55.64 increasing to 37.50 and 83.14, respectively).
This may be due to architectural differences: the MLP of LLaVA-NeXT-Vicuna handles larger token sizes efficiently, whereas the Q-former of InternLM-XComposer struggles with increased computational complexity and information focus.

\textbf{Free-prompt Following Ability.} From Table~\ref{tab:results}, we can see that our model exhibits strong free-prompt following ability, with FPF ranging from 86.42 to 91.00, attributed to our free-prompt construction pipeline. High-quality free-prompt and draft pairs significantly enhance output controllability. 
As for the specific dimensions, Fig.~\ref{fig:free_prompt_following} shows that LLaVA-NeXT-Mistral and LLaVA-NeXT-Qwen excel in free-prompt following across categories like Visual Storyline, Music, and TTS Timbre, with high scores of 16.47 and 16.40, 9.76 and 9.77, and 9.52 and 9.50, respectively. In contrast, MiniCPMV2.6 performs weaker, particularly in Visual Storyline (15.63) and Duration (9.21).

\textbf{Script Quality \& Decoration Evaluation.} Fig.~\ref{fig:first_figure} shows the quality of voice-over script perfectly meets the requirements of advertising narrative videos, with overall SQ scores ranging from 70.14 to 78.29 in Table~\ref{tab:results}. For the detailed aspects, Fig.~\ref{fig:script_quality} reveals that LLaVA-NeXT-Qwen consistently performs well across all categories, including Basic, Native Language \& Tone, Touch the Audience, and Creative Narrative.
Table~\ref{tab:results} shows the model has high accuracy in recommending tags for decorative elements, with DTPR ranging from 77.40/77.02 to 80.54/80.28. Effectively coordinating multiple decorative elements is crucial in production, and our model excels at this (see cases in Appendix~\ref{app-cases}). 
Table~\ref{tab:decorative_tags} shows that model performance across different components (TTS Timbre, Avatar, Music). LLaVA-NeXT-Vicuna excels in most metrics. However, nearly all models have lower precision and recall for avatar, due to the higher complexity of avatar tags.

\subsection{Human Evaluation Results.} We have additionally executed human evaluations, categorized into five distinct components: Basic, Visual, Script, Voice (TTS), and Music. 
Each component is further subdivided into quality evaluation and follow-up ability assessment. The quality evaluation employs a 5-point scale, whereas the follow-up ability assessment utilizes a 3-point scale, with elevated scores signifying superior performance (see detailed criteria in Appendix~\ref{app-humaneval}). 
As shown in Table~\ref{tab:human}, our finetuned model surpasses GPT-4o in both quality and follow-up ability, particularly demonstrating improvements in the Basic, Script, and Visual aspects with gains of 0.32, 0.43, and 0.38, respectively, in InternLM-XComposer.

\subsection{Transferability of Task in Shot2story.} Table \ref{tab:story} evaluates the performance of LLaVA-NeXT-Vicuna and InternLM-XComposer on the Shot2story dataset. Through CRA and CSA, we can verify the efficacy of the slow-fast strategy. However, the slow-fast model of InternLM-XComposer performs less effectively than the single fast pathway model in generating shot captions. We attribute this to the large number of visual token inputs, which may negatively impact the text generation ability for this task.
However, for the task on this dataset, we visualized the inference result and found that our model can simultaneously and excellently complete shots ranking, selection, and caption generation (see the Appendix~\ref{app-shot2story} for details), validating our method's applicability to other editing scenarios.

\begin{table}[t]
    \centering
    \resizebox{\columnwidth}{!}{
        \begin{tabular}{l|cc|cc|cc|cc}
            \toprule
            Model & \multicolumn{2}{c|}{TTS Timbre} & \multicolumn{2}{c|}{Avatar} & \multicolumn{2}{c|}{Music} &  \multicolumn{2}{c}{Average} \\
            \cline{2-3} \cline{4-5} \cline{6-7} \cline{8-9} 
            - & Precision & Recall & Precision & Recall & Precision & Recall & Precision & Recall\\
            \midrule
            \rowcolor[gray]{0.93}  
            LLaVA-NeXT-Vicuna & 88.98 & \textbf{93.65} & \textbf{73.65} & 65.87 & \textbf{78.98} & \textbf{81.33} & \textbf{80.54} & \textbf{80.28} \\
            LLaVA-NeXT-Mistral & 89.03 & 93.36 & 69.62 & 66.14 & 78.1 & 79.66 & 78.92 & 79.72 \\
            LLaVA-NeXT-Qwen & 89.16 & 93.07 & 67.51 & \textbf{68.11} & 77.69 & 79.48 & 78.12 & 80.22 \\
            InternLM-XComposer & 88.26 & 91.29 & 67.99 & 65.96 & 77.4 & 78.34 & 77.88 & 78.53 \\
            InternVL & \textbf{89.18} & 92.98 & 71.24 & 66.47 & 78.42 & 80.3 & 79.61 & 79.92 \\
            MiniCPMV2.6 & 87.46 & 89.16 & 68.26 & 64.52 & 76.47 & 77.37 & 77.40 & 77.02 \\
            \bottomrule
        \end{tabular}
    }
    \caption{Details of various models on decorative tags generation (Qwen, fps/token:2/9, others, fps/token:2/4).}
    \label{tab:decorative_tags}
    \vspace{-0.8em}
\end{table}

\begin{table}[t]
    \centering
    \resizebox{\columnwidth}{!}{
        \begin{tabular}{lcccccccccc}
            \toprule
            Model & Basic & \makecell{Basic\\(FU)} & Script & \makecell{Script\\(FU)} & Visual & \makecell{Visual\\(FU)} & Voice & \makecell{Voice\\(FU)} & Music & \makecell{Music\\(FU)} \\

            \midrule
            GPT-4o & 3.55 & 1.70 & 3.55 & 1.53 & 3.68 & 1.59 & 4.40 & 1.88 & 4.40 & 1.81 \\
            \rowcolor[gray]{0.93}  
            LLaVA-NeXT-Vicuna & 3.81 & \textbf{1.83} & 3.78 & 1.64 & 3.75 & 1.71 & \textbf{4.59} & 1.85 & \textbf{4.52} & \textbf{1.82} \\
            \rowcolor[gray]{0.93}  
            InternLM-XComposer & \textbf{3.87} & 1.76 & \textbf{3.98} & \textbf{1.75} & \textbf{4.06} & \textbf{1.77} & \textbf{4.59} & \textbf{1.89} & \textbf{4.52} & \textbf{1.82} \\
            \bottomrule
        \end{tabular}
    }
    \caption{Human evaluation of GPT-4o, LLaVA-NeXT-Vicuna, and InternLM-XComposer (fps/token:0.125/64, FU is Follow-up).}
    \label{tab:human}
    \vspace{-0.8em}
\end{table}

\begin{table}[t]
    \centering
    \resizebox{\columnwidth}{!}{
        \begin{tabular}{lccccccc}
            \toprule
            Model & Setting(fps/token)& CRA & CSA & BLEU & ROUGE-L & CIDEr & METEOR  \\

            \midrule
            LLaVA-NeXT-Vicuna &2/4 & 12.64 & 66.73  & 10.18 & 27.98 & 27.20 & 31.66  \\
            LLaVA-NeXT-Vicuna &0.125/64 & 12.15 & 72.07 & 9.91 & 27.37 & 25.72 & 30.94  \\
            \rowcolor[gray]{0.93}  
            LLaVA-NeXT-Vicuna &\textit {fast:}2/4 \textit {slow:}0.125/64 & \textbf{15.21} & \textbf{78.26} & \textbf{10.87} & \textbf{28.87} & \textbf{31.92} & \textbf{32.69}  \\
            \midrule
            InternLM-XComposer &2/4 & 20.86 & 75.42 & \textbf{10.43} & \textbf{28.67} & \textbf{30.05} & \textbf{33.57}  \\            
            InternLM-XComposer &0.125/64 & 15.03 & 77.99 & 9.22 & 26.54 & 22.96 & 30.87  \\
            \rowcolor[gray]{0.93}  
            InternLM-XComposer &\textit {fast:}2/4 \textit {slow:}0.125/64 & \textbf{21.05} & \textbf{80.08} & 9.84 & 27.92 & 27.70 & 32.52  \\
            \bottomrule
        \end{tabular}
    }
    \caption{Evaluating performance on Shot2story dataset.}
    \label{tab:story}
    \vspace{-1.2em}
\end{table}

%% file: sec/5-conclusion.tex
\section{Conclusion}
We developed a unified video editing model for advertising marketing, leveraging multimodal Large Language Models. The model requires only product information and specific video requirements to autonomously generate a video editing draft. To handle more information and balance temporal and spatial dimensions, we compress visual token sizes and use a denser slow-fast approach to process frames effectively. Additionally, we implemented a free-prompt strategy, allowing users to specify textual requirements, enhancing output precision for practical business applications. Our model also adapts easily to other editing scenarios with minimal modifications.

%% file: sec/X_suppl.tex
\clearpage
\setcounter{page}{1}
\maketitlesupplementary

\section{Appendix}
\label{sec:appendix}
\subsection{Data Construction Templates}
\label{app-data}
As illustrated in Fig.~\ref{fig:template}, this template is designed for data construction, offering a detailed example of our input-output structure. 
The process involves taking product information, materials (video clips) information, and specific requirements (free prompt) as input.
For the details of the video clips, each clip includes its index, duration, and placeholders (like "<image>") for frames corresponding to different frame-rate pathways. 
Before entering the LLM, these placeholders are replaced with the respective visual tokens.
Regarding the output, the model generates a video draft in JSON format, which includes the "voice\_over\_track", "video\_nodes\_track", and "decoration\_setting". 
In the "voice\_over\_track", it specifies the start and end times ("target\_start", "target\_end") for each voice-over sentence appearance in the video. 
The "video\_nodes\_track" includes the sequence of selected clips, where each clip comprises its index, start and end time of appearance ("target\_start", "target\_end"), and the start time of playback in the original clip ("source\_start").
Through the above simple formulation, we can clearly define the input and output of the model.

\subsection{Post-Processing and Transformation}
\label{app-draft}
\begin{figure}[h]
    \centering
    \includegraphics[width=0.95\columnwidth]{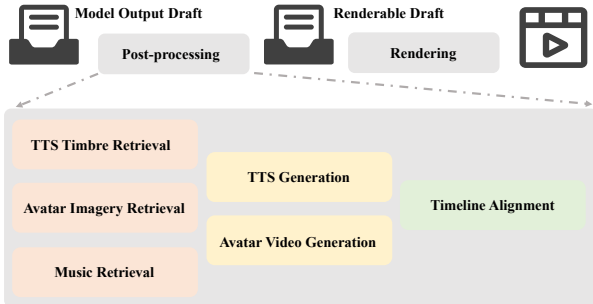} 
    \caption{Post-processing and rendering process.}
    \label{fig:postprocessing}
\end{figure}
Fig.~\ref{fig:postprocessing} depicts the post-processing of the model-generated draft and the subsequent rendering process into a final video. 
Initially, the draft generated by the model is parsed. We use the TTS timbre tags and avatar settings output by the model to match the corresponding voice-over timbre and the avatar's appearance. 
Simultaneously, the music tags are used to retrieve background music of the appropriate style.
Next, we transform the voice-over content into TTS and submit the avatar rendering task to create the avatar narration video.
We then adjust the video clips in the video nodes track to ensure alignment with the generated TTS and avatar video timelines.
Finally, a new renderable video draft is generated and submitted for final rendering, producing the fully rendered video.

\begin{figure*}[t]
    \centering
    \includegraphics[width=\textwidth]{./figure/template.pdf} 
    \caption{Details of data sample construction template.}
    \label{fig:template}
\end{figure*}

\subsection{Prompts for GPT-4o Evaluation}
\label{app-prompts}



As mentioned in Section~\ref{metrics}, the evaluations for free-prompt following capability and script quality are documented in the files "free\_prompt\_gpt4o\_eval.txt" and "script\_quality\_gpt4o\_eval.txt", respectively.
For further details, please refer to the corresponding documents.
Specifically, for the free-prompt evaluation, we use a total score of 100 points, distributed across various dimensions: Video Duration (10), Visual Storyline (20), Target Audience (10), Script Routine (10), Selling-points (20), Avatar (10), TTS Timbre (10), and Music (10).
Similarly, for the script quality evaluation, the total score is 100 points, divided into the following categories: Basic (30), Native Language \& Tone (15), Touch the Audience (15), and Creative Narrative (40).
Notably, within the script evaluation, we have included both good and bad examples to ensure that GPT-4o comprehends the specific details.

\subsection{LLM Metrics for Draft Evaluation}
\label{app-llm}
As mentioned in Section~\ref{metrics}, we present the NLP metrics in Table~\ref{tab:llm_metrics}.
Notably, without fine-tuning, GPT-4o~\cite{achiam2023gpt} significantly underperforms compared to LLaVA-NeXT-Vicuna~\cite{li2024llavanext-interleave} and InternLM-XComposer~\cite{zhang2023internlm} at the same fps settings (0.125fps). 
However, there is not much difference in these indicators in the denser, slow-fast strategy. 
This is because these indicators are used for machine translation and the draft generation task contains many common structured words, resulting in a relatively low proportion of truly meaningful words. Therefore, using typical NLP metrics for evaluation is not feasible.

\begin{table}[t]
    \centering
    \resizebox{\columnwidth}{!}{
        \begin{tabular}{lccccc}
            \toprule
            Model & Setting(fps/token)& BLEU & ROUGE-L & CIDEr & METEOR  \\
            \midrule
            GPT-4o & 0.125/- & 21.24 & 43.99 & 8.70 & 50.27 \\
            \midrule
            \multicolumn{6}{l}{\textit {\textbf{Denser Frame}}} \\
            \midrule
            LLaVA-NeXT-Vicuna-7B & 0.125/1 & 26.93 & 48.75 & 33.40 & 63.28 \\
            LLaVA-NeXT-Vicuna-7B & 2/1 & 26.91 & 48.64 & 32.18 & 63.43 \\
            LLaVA-NeXT-Vicuna-7B & 0.125/4 & 26.87 & 48.80 & 32.97 & 63.51 \\
            LLaVA-NeXT-Vicuna-7B & 2/4 & 28.01 & 49.23 & 32.80 & 63.55 \\
            InternLM-XComposer-7B & 0.125/1 & 23.45 & 46.43 & 27.47 & 63.10 \\
            InternLM-XComposer-7B & 2/1 & 23.60 & 46.59 & 29.12 & 63.35 \\
            InternLM-XComposer-7B & 0.125/4 & 23.51 & 46.60 & 27.55 & 63.70 \\
            InternLM-XComposer-7B & 2/4 & 23.15 & 46.25 & 27.56 & 62.86 \\
            LLaVA-NeXT-Mistral-7B & 2/1 & 27.51 & 49.25 & 36.37 & 63.18 \\
            LLaVA-NeXT-Mistral-7B & 2/4 & 29.36 & 50.34 & 41.32 & 63.84 \\
            LLaVA-NeXT-Qwen-7B & 2/1 & 26.55 & 48.50 & 32.99 & 62.11 \\
            LLaVA-NeXT-Qwen-7B & 2/4 & 28.98 & 49.58 & 33.58 & 62.44 \\
            InternVL-26B & 2/1 & 23.84 & 47.08 & 30.17 & 64.01 \\
            InternVL-26B & 2/4 & 24.02 & 47.19 & 31.10 & 64.09 \\
            MiniCPMV2.6-7B & 2/1 & 26.72 & 47.46 & 32.32 & 60.01 \\
            MiniCPMV2.6-7B & 2/4 & 26.05 & 46.90 & 31.55 & 59.33 \\
            \midrule
            \multicolumn{6}{l}{\textit {\textbf{Slow-fast Strategy}}} \\
            \midrule
            LLaVA-NeXT-Vicuna & \textit {fast:}2/4 & 28.01 & 49.23 & 32.80 & 63.55 \\
            LLaVA-NeXT-Vicuna & \textit {slow:}0.5/16 & 28.77 & 49.53 & 33.29 & 63.97  \\
            LLaVA-NeXT-Vicuna & \textit {slow:}0.125/64 & 28.95 & 49.72 & 33.63 & 63.91 \\
            LLaVA-NeXT-Vicuna & \textit {fast:}2/4 \textit {slow:}0.5/16 & 28.71 & 49.45 & 34.66 & 63.79  \\
            LLaVA-NeXT-Vicuna & \textit {fast:}2/4 \textit {slow:}0.125/64 & 26.74 & 48.38 & 33.10 & 61.90  \\
            \midrule
            InternLM-XComposer & \textit {fast:}2/4 & 23.15 & 46.25 & 27.56 & 62.86 \\          
            InternLM-XComposer & \textit {slow:}0.5/16 & 23.11 & 46.32 & 26.83 & 63.08  \\
            InternLM-XComposer & \textit {slow:}0.125/64 & 23.51 & 46.60 & 27.55 & 63.69  \\
            InternLM-XComposer & \textit {fast:}2/4 \textit {slow:}0.5/16 & 22.52 & 45.80 & 27.81 & 62.66  \\
            InternLM-XComposer & \textit {fast:}2/4 \textit {slow:}0.125/64 & 22.52 & 46.17 & 28.00 & 62.85  \\
            \bottomrule
        \end{tabular}
    }
    \caption{Details of the NLP metrics results.}
    \label{tab:llm_metrics}
    \vspace{-1.2em}
\end{table}

\subsection{GPT-4o Draft Generation Details}
\label{app-gpt4pipe}
To validate that our proposed method surpasses the GPT after fine-tuning, we developed a draft generation scheme based on GPT-4o, ensuring identical input and output conditions. 
The prompt used is detailed in the file "gpt4o\_gen\_draft\_prompt.txt". 
Due to the limitation on the number of input images for GPT-4o, we utilized an input rate of only 0.125 fps. 
Furthermore, to enhance the consistency of the model's output format, we incorporated few-shot examples as references within the prompt.

\subsection{Human Evaluation Criteria}
\label{app-humaneval}
For the human evaluation, we recruited three experts from the video advertising production industry to assess generated videos by different models. 
As seen in Table~\ref{tab:human}, the average score from the three evaluators is used as the manual evaluation result. 
Additionally, the detailed evaluation criteria is provided in Table~\ref{tab:human_eval}.

\begin{table*}[t]
    \vspace{0.5em}
    \centering
    \begin{tabular}{ p{0.1\textwidth} | p{0.1\textwidth} | p{0.3\textwidth} | p{0.4\textwidth} }
    \toprule
    \textbf{Dimension} & \textbf{Quality \& Follow-up} &\textbf{Description} & \textbf{Scoring Rules} (Quality total score: 5, Follow-up total score: 3) \\
    \midrule
    \textbf{Basic} & Quality Evaluation & Evaluate the video based on the reviewer’s willingness to watch or purchase and the effectiveness of the script, visual, and auditory elements. & 
        \begin{itemize}
            \vspace{-0.8em}
            \item 4-5: Highly likely to purchase after watching.
            \item 3: Would buy if needed, nothing special.
            \item 1-2: Low purchase intent, unclear advantages.
            \vspace{-0.8em}
        \end{itemize} \\
    \cline{2-4}
     & Follow-up Assessment & Evaluate the video duration, target audience, and overall video content for consistency with the free-prompt. & 
        \begin{itemize}
            \vspace{-0.8em}
            \item 3: The overall content is highly consistent with free-prompt.
            \item 2: Part of them are followed.
            \item 1: None of them are followed.
            \vspace{-0.8em}
        \end{itemize} \\
    \midrule
    \textbf{Script} & Quality Evaluation & Evaluate the script on hook appeal, Call to Action (CTA) effectiveness, core selling-points or practical benefits, and incorporation of personal experiences. & 
        \begin{itemize}
            \vspace{-0.8em}
            \item 4-5: Perfect script, engaging and product selling points clearly conveyed.
            \item 3: Average, no major issues but lacks highlights.
            \item 1-2: Clear flaws or poor effect, fails to convey product value.
            \vspace{-0.8em}
        \end{itemize} \\
    \cline{2-4}
     & Follow-up Assessment & Evaluate whether the script follows the free prompt in terms of routine structure, emphasized selling points, and accuracy of price or brand. & 
        \begin{itemize}
            \vspace{-0.8em}
            \item 3: Follows all script routine, selling points, and other mentioned elements.
            \item 2: Part of them are followed.
            \item 1: None of them are followed.
            \vspace{-0.8em}
        \end{itemize} \\
    \midrule
    \textbf{Visual} & Quality Evaluation & Evaluate the visual quality of the video, including the appropriateness of digital avatar, logical video clip editing, and consistency with script content. & 
        \begin{itemize}
            \vspace{-0.8em}
            \item 4-5: Perfect logic and editing of clips.
            \item 3: No significant issues with clips use and matching.
            \item 1-2: Clear errors, significant mismatches, or illogical editing.
            \vspace{-0.8em}
        \end{itemize} \\
    \cline{2-4}
     & Follow-up Assessment & Evaluate whether the visual storyline and the setting of the avatar are consistent with the free-prompt. & 
        \begin{itemize}
            \vspace{-0.8em}
            \item 3: The visual content and avatar completely conform to the free-prompt.
            \item 2: Part of them are followed.
            \item 1: None of them are followed.
            \vspace{-0.8em}
        \end{itemize} \\
    \midrule
    \textbf{Voice} & Quality Evaluation & Evaluate the authenticity, naturalness, tone, and emotional impact of the quality of TTS, and its alignment with the script. & 
        \begin{itemize}
            \vspace{-0.8em}
            \item 4-5: Accurate timbre, accent, and tone, and the voice is positive and engaging.
            \item 3: No major issues, but lacks highlights.
            \item 1-2: Significant flaws, gender or accent errors.
            \vspace{-0.8em}
        \end{itemize} \\
    \cline{2-4}
     & Follow-up Assessment & Evaluate if the voice's timbre (age, gender) and accent match the free prompt. &
        \begin{itemize}
            \vspace{-0.8em}
            \item 3: Timbre and accent completely match the free-prompt.
            \item 2: Part of them are followed.
            \item 1: None of them are followed.
            \vspace{-0.8em}
        \end{itemize} \\
    \midrule
    \textbf{Music} & Quality Evaluation & Evaluate whether the background music is appropriate and engaging. & 
        \begin{itemize}
            \vspace{-0.8em}
            \item 4-5: Music is highly relevant or perfectly matches the product.
            \item 3: Music generally matches the product.
            \item 1-2: Clear issue, the music not being well-matched.
            \vspace{-0.8em}
        \end{itemize} \\
    \cline{2-4}
     & Follow-up Assessment & Evaluate the consistency of the music's genre and emotion with the free-prompt requirements. & 
        \begin{itemize}
            \vspace{-0.8em}
            \item 3: Music perfectly matches with free-prompt.
            \item 2: Part of them are followed.
            \item 1: None of them are followed.
            \vspace{-0.8em}
        \end{itemize} \\
    \bottomrule
    \end{tabular}
    \caption{Details of human evaluation criteria.}
    \label{tab:human_eval}
\end{table*}

\subsection{Case Studies for Advertising Videos}
\label{app-cases}

\begin{figure*}[t]
    \centering
    \includegraphics[width=0.95\textwidth]{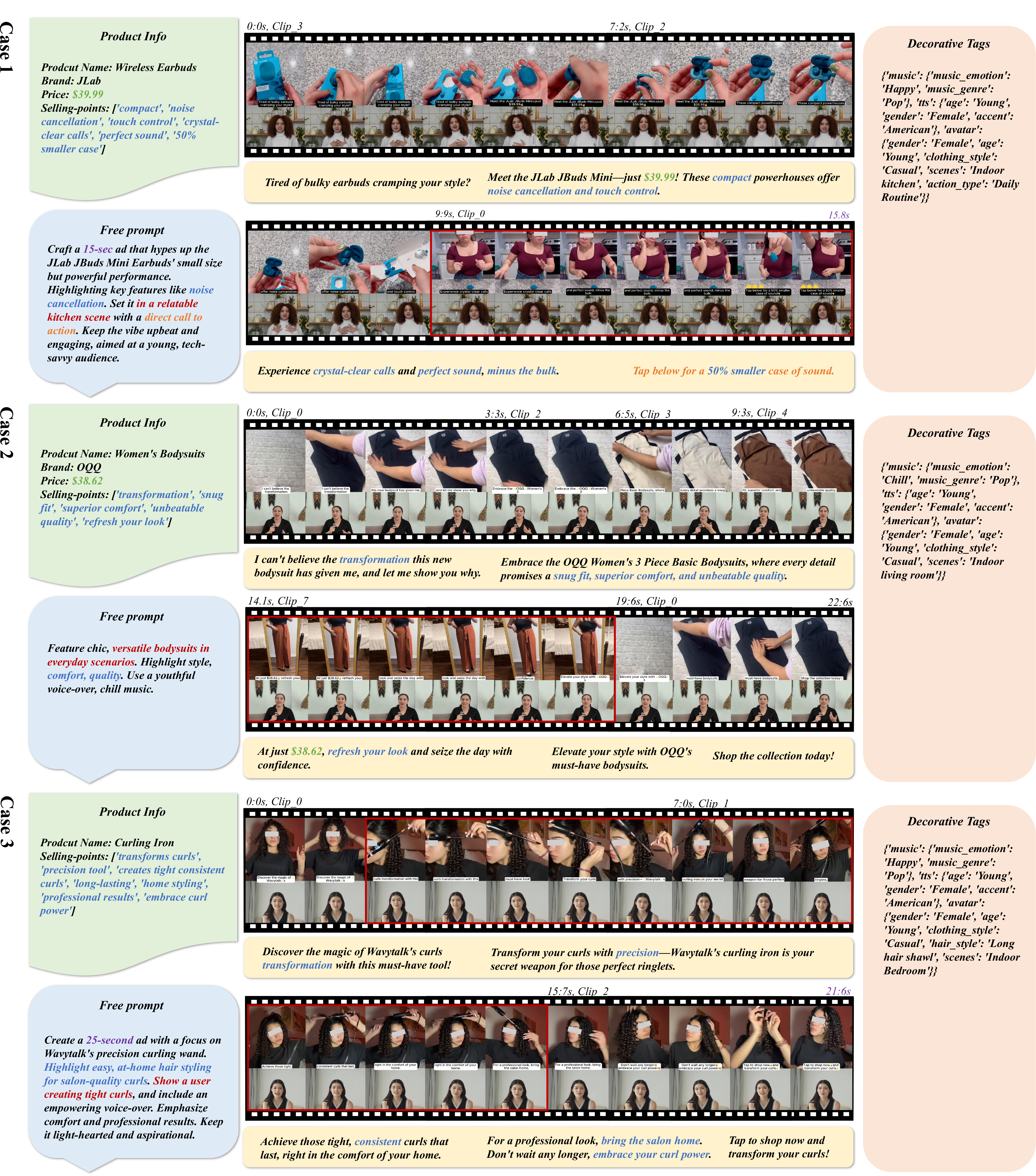} 
    \caption{More examples on VideoAds dataset.}
    \label{fig:showcase2}
    \vspace{-0.8em}
\end{figure*}

In Fig.~\ref{fig:showcase2}, we showcase additional videos generated by our method, demonstrating that our model consistently produces high-quality and controllable outputs. 
Furthermore, we provide the set of decorative element tags recommended by the model. 
For example, in case 1, our model recommends a "young female avatar" in a "kitchen setting", which matches the environment in the original video clips. 
This scene is enhanced by "popular" background music, appealing to the youthful audience targeted by the earbuds being sold. 
It is clear that the TTS timbre, avatar settings, and background music chosen by the model are seamlessly coordinated and align perfectly with the brand's identity and target audience.

\subsection{Case Studies for Shot2story Dataset}
\label{app-shot2story}
\vspace{-0.5em}

\begin{figure*}[t]
    \centering
    \includegraphics[width=0.95\textwidth]{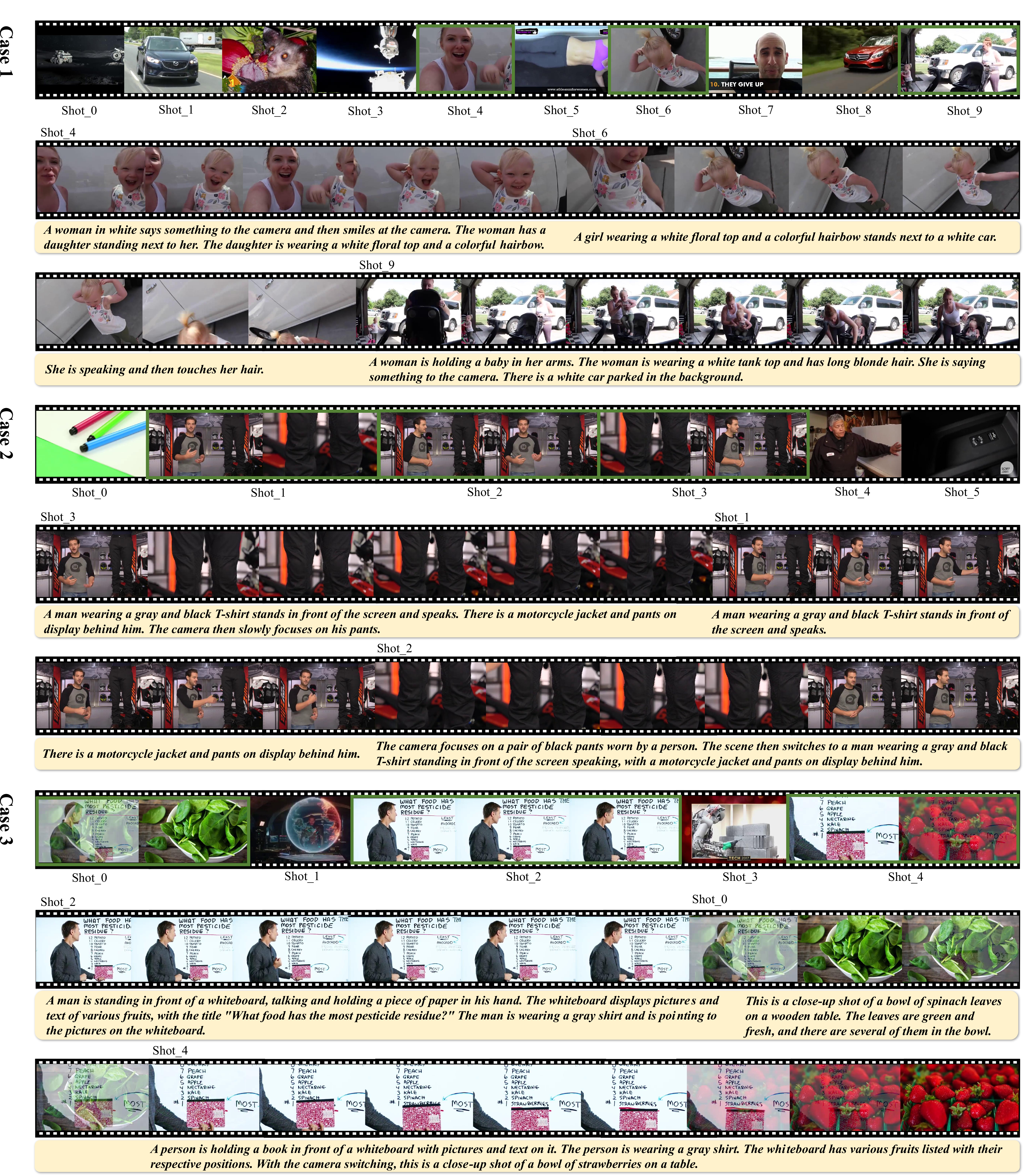} 
    \caption{Examples on Shot2story dataset.}
    \label{fig:shot2story}
    \vspace{-0.8em}
\end{figure*}

As shown in Fig~\ref{fig:shot2story}, we define a new editing scenario on Shot2story~\cite{han2023shot2story20k} dataset. 
The model's task is to identify all the shots that can be combined into a coherent storyline from a series of video shots, correctly sort these shots, and generate a caption for each one.
Due to our denser slow-fast strategy, the model captures motion within the shots comprehensively. 
For example, in the caption of shot\_3 in case 2, the model perceives the movement of the camera and naturally transitions from "a man" to "pants". Similarly, in shot\_4 of case 3, it detects the scene switching from "the text on the whiteboard" to "strawberries".

\subsection{Details of Decoration Tags}
\label{app-tags}

Table~\ref{decorative_tags_details} presents the labels of decorative elements, which are classed into three main categories: TTS, Avatar, and Music, containing 3, 14, and 2 subcategories respectively, accounting for a total of 98 labels. Notably, the Avatar category, having the highest number of labels, exhibits lower precision and recall as seen in Table~\ref{tab:decorative_tags}.
In practical applications, we enhance and enrich the final video quality and effect by retrieving the corresponding decorative elements based on the tags recommended by the model.

\begin{table*}[h]
    \centering
    \begin{tabular}{ p{0.2\textwidth} | p{0.3\textwidth} | p{0.4\textwidth} }
    \toprule
    \textbf{Category} & \textbf{Subcategory} & \textbf{Labels} \\
    \midrule
    \textbf{TTS} & Age & Young, Middle Aged, Old \\
    \cline{2-3}
     & Gender & Female, Male \\
    \cline{2-3}
     & Accent & American, African American, Southeast Asian, British, European, Indian, Australian \\
    \midrule
    
    \textbf{Avatar} & Race & White, Black, Southeastern Asian, Latino, East Asian \\
    \cline{2-3}
     & Gender & Female, Male \\
    \cline{2-3}
     & Age & Young, Middle Aged, Old \\
    \cline{2-3}
     & Body Shape & Average, Large Size \\
    \cline{2-3}
     & Body Gesture & Standing frontally, Sit upright, Sitting sideways, Stand sideways, Taking a selfie walking \\
    \cline{2-3}
     & Hand Gesture & Handheld product, Handheld microphone, Handheld mobile phone, Play on the phone \\
    \cline{2-3}
     & Scenes & Indoor living room, In car, Indoor kitchen, Street, No background, In the office, Garden, Broadcast, Solid color background, E-sports room \\
    \cline{2-3}
     & Clothing Style & Casual, Sporty, Formal, Traditional, Funny \\
    \cline{2-3}
     & Clothing Color System & Light color, Deep color, Multi color mix \\
    \cline{2-3}
     & Hair Style & Long hair tied, Short hair, Long hair shawl \\
    \cline{2-3}
     & Hair Color & Blonde, Brown, Black, Grey, White, Red \\
    \cline{2-3}
     & Beard & No beard, Light beard, Heavy beard \\
    \cline{2-3}
     & Emotions & Satisfied, Enlightened, Welcoming, Relieved, Dissatisfied, Surprised, Anticipated, Curious, Troubled \\
    \cline{2-3}
     & Action Type & Daily Routine, Special \\
    \midrule
    
    \textbf{Music} & Music Emotion & Happy, Romantic, Chill, Dynamic, Weird, Cute, Excited, Tense, Sorrow, Angry \\
    \cline{2-3}
     & Music Genre & Pop, Easy listening, Hip Hop/Rap, New Age, Blues, Country, Metal, Electronic, Rock, Latin, Experimental, R\&B/Soul, Jazz, Classical \\
    \bottomrule
    \end{tabular}
    \caption{Details of decorative element labels}
    \label{decorative_tags_details}
\end{table*}





%% file: main.bbl
\begin{thebibliography}{50}
\providecommand{\natexlab}[1]{#1}
\providecommand{\url}[1]{\texttt{#1}}
\expandafter\ifx\csname urlstyle\endcsname\relax
  \providecommand{\doi}[1]{doi: #1}\else
  \providecommand{\doi}{doi: \begingroup \urlstyle{rm}\Url}\fi

\bibitem[Achiam et~al.(2023)Achiam, Adler, Agarwal, Ahmad, Akkaya, Aleman, Almeida, Altenschmidt, Altman, Anadkat, et~al.]{achiam2023gpt}
Josh Achiam, Steven Adler, Sandhini Agarwal, Lama Ahmad, Ilge Akkaya, Florencia~Leoni Aleman, Diogo Almeida, Janko Altenschmidt, Sam Altman, Shyamal Anadkat, et~al.
\newblock Gpt-4 technical report.
\newblock \emph{arXiv preprint arXiv:2303.08774}, 2023.

\bibitem[Ahanger and Little(1998)]{ahanger1998automatic}
Gulrukh Ahanger and Thomas~DC Little.
\newblock Automatic composition techniques for video production.
\newblock \emph{IEEE Transactions on Knowledge and Data Engineering}, 10\penalty0 (6):\penalty0 967--987, 1998.

\bibitem[Alayrac et~al.(2022)Alayrac, Donahue, Luc, Miech, Barr, Hasson, Lenc, Mensch, Millican, Reynolds, et~al.]{alayrac2022flamingo}
Jean-Baptiste Alayrac, Jeff Donahue, Pauline Luc, Antoine Miech, Iain Barr, Yana Hasson, Karel Lenc, Arthur Mensch, Katherine Millican, Malcolm Reynolds, et~al.
\newblock Flamingo: a visual language model for few-shot learning.
\newblock \emph{Advances in neural information processing systems}, 35:\penalty0 23716--23736, 2022.

\bibitem[Arev et~al.(2014)Arev, Park, Sheikh, Hodgins, and Shamir]{arev2014automatic}
Ido Arev, Hyun~Soo Park, Yaser Sheikh, Jessica Hodgins, and Ariel Shamir.
\newblock Automatic editing of footage from multiple social cameras.
\newblock \emph{ACM Transactions on Graphics (TOG)}, 33\penalty0 (4):\penalty0 1--11, 2014.

\bibitem[Argaw et~al.(2022)Argaw, Heilbron, Lee, Woodson, and Kweon]{argaw2022anatomy}
Dawit~Mureja Argaw, Fabian~Caba Heilbron, Joon-Young Lee, Markus Woodson, and In~So Kweon.
\newblock The anatomy of video editing: A dataset and benchmark suite for ai-assisted video editing.
\newblock In \emph{European Conference on Computer Vision}, pages 201--218. Springer, 2022.

\bibitem[Bai et~al.(2023)Bai, Bai, Chu, Cui, Dang, Deng, Fan, Ge, Han, Huang, et~al.]{bai2023qwen}
Jinze Bai, Shuai Bai, Yunfei Chu, Zeyu Cui, Kai Dang, Xiaodong Deng, Yang Fan, Wenbin Ge, Yu Han, Fei Huang, et~al.
\newblock Qwen technical report.
\newblock \emph{arXiv preprint arXiv:2309.16609}, 2023.

\bibitem[Banerjee and Lavie(2005)]{banerjee2005meteor}
Satanjeev Banerjee and Alon Lavie.
\newblock Meteor: An automatic metric for mt evaluation with improved correlation with human judgments.
\newblock In \emph{Proceedings of the acl workshop on intrinsic and extrinsic evaluation measures for machine translation and/or summarization}, pages 65--72, 2005.

\bibitem[Cai et~al.(2024)Cai, Cao, Chen, Chen, Chen, Chen, Chen, Chen, Chen, Chu, et~al.]{cai2024internlm2}
Zheng Cai, Maosong Cao, Haojiong Chen, Kai Chen, Keyu Chen, Xin Chen, Xun Chen, Zehui Chen, Zhi Chen, Pei Chu, et~al.
\newblock Internlm2 technical report.
\newblock \emph{arXiv preprint arXiv:2403.17297}, 2024.

\bibitem[Chen et~al.(2024)Chen, Wu, Wang, Su, Chen, Xing, Zhong, Zhang, Zhu, Lu, et~al.]{chen2024internvl}
Zhe Chen, Jiannan Wu, Wenhai Wang, Weijie Su, Guo Chen, Sen Xing, Muyan Zhong, Qinglong Zhang, Xizhou Zhu, Lewei Lu, et~al.
\newblock Internvl: Scaling up vision foundation models and aligning for generic visual-linguistic tasks.
\newblock In \emph{Proceedings of the IEEE/CVF Conference on Computer Vision and Pattern Recognition}, pages 24185--24198, 2024.

\bibitem[Chiang et~al.(2023)Chiang, Li, Lin, Sheng, Wu, Zhang, Zheng, Zhuang, Zhuang, Gonzalez, et~al.]{chiang2023vicuna}
Wei-Lin Chiang, Zhuohan Li, Zi Lin, Ying Sheng, Zhanghao Wu, Hao Zhang, Lianmin Zheng, Siyuan Zhuang, Yonghao Zhuang, Joseph~E Gonzalez, et~al.
\newblock Vicuna: An open-source chatbot impressing gpt-4 with 90\%* chatgpt quality.
\newblock \emph{See https://vicuna. lmsys. org (accessed 14 April 2023)}, 2\penalty0 (3):\penalty0 6, 2023.

\bibitem[Chua and Ruan(1995)]{chua1995video}
Tat-Seng Chua and Li-Qun Ruan.
\newblock A video retrieval and sequencing system.
\newblock \emph{ACM Transactions on Information Systems (TOIS)}, 13\penalty0 (4):\penalty0 373--407, 1995.

\bibitem[Dong et~al.(2024)Dong, Zhang, Zang, Cao, Wang, Ouyang, Wei, Zhang, Duan, Cao, Zhang, Li, Yan, Gao, Zhang, Li, Li, Chen, He, Zhang, Qiao, Lin, and Wang]{internlmxcomposer2}
Xiaoyi Dong, Pan Zhang, Yuhang Zang, Yuhang Cao, Bin Wang, Linke Ouyang, Xilin Wei, Songyang Zhang, Haodong Duan, Maosong Cao, Wenwei Zhang, Yining Li, Hang Yan, Yang Gao, Xinyue Zhang, Wei Li, Jingwen Li, Kai Chen, Conghui He, Xingcheng Zhang, Yu Qiao, Dahua Lin, and Jiaqi Wang.
\newblock Internlm-xcomposer2: Mastering free-form text-image composition and comprehension in vision-language large model.
\newblock \emph{arXiv preprint arXiv:2401.16420}, 2024.

\bibitem[Feichtenhofer et~al.(2019)Feichtenhofer, Fan, Malik, and He]{feichtenhofer2019slowfast}
Christoph Feichtenhofer, Haoqi Fan, Jitendra Malik, and Kaiming He.
\newblock Slowfast networks for video recognition.
\newblock In \emph{Proceedings of the IEEE/CVF international conference on computer vision}, pages 6202--6211, 2019.

\bibitem[Han et~al.(2023)Han, Yang, Chang, and Wang]{han2023shot2story20k}
Mingfei Han, Linjie Yang, Xiaojun Chang, and Heng Wang.
\newblock Shot2story20k: A new benchmark for comprehensive understanding of multi-shot videos.
\newblock \emph{arXiv preprint arXiv:2312.10300}, 2023.

\bibitem[Huang et~al.(2024)Huang, Liao, Radhakrishnan, Yin, Molchanov, Yu, and Kautz]{huang2024lita}
De-An Huang, Shijia Liao, Subhashree Radhakrishnan, Hongxu Yin, Pavlo Molchanov, Zhiding Yu, and Jan Kautz.
\newblock Lita: Language instructed temporal-localization assistant.
\newblock \emph{arXiv preprint arXiv:2403.19046}, 2024.

\bibitem[Huber et~al.(2019)Huber, Shin, Russell, Wang, and Mysore]{huber2019b}
Bernd Huber, Hijung~Valentina Shin, Bryan Russell, Oliver Wang, and Gautham~J Mysore.
\newblock B-script: Transcript-based b-roll video editing with recommendations.
\newblock In \emph{Proceedings of the 2019 CHI Conference on Human Factors in Computing Systems}, pages 1--11, 2019.

\bibitem[Jiang et~al.(2023)Jiang, Sablayrolles, Mensch, Bamford, Chaplot, Casas, Bressand, Lengyel, Lample, Saulnier, et~al.]{jiang2023mistral}
Albert~Q Jiang, Alexandre Sablayrolles, Arthur Mensch, Chris Bamford, Devendra~Singh Chaplot, Diego de~las Casas, Florian Bressand, Gianna Lengyel, Guillaume Lample, Lucile Saulnier, et~al.
\newblock Mistral 7b.
\newblock \emph{arXiv preprint arXiv:2310.06825}, 2023.

\bibitem[Leake and Li(2024)]{leake2024chunkyedit}
Mackenzie Leake and Wilmot Li.
\newblock Chunkyedit: Text-first video interview editing via chunking.
\newblock In \emph{Proceedings of the CHI Conference on Human Factors in Computing Systems}, pages 1--16, 2024.

\bibitem[Leake et~al.(2017)Leake, Davis, Truong, and Agrawala]{leake2017computational}
Mackenzie Leake, Abe Davis, Anh Truong, and Maneesh Agrawala.
\newblock Computational video editing for dialogue-driven scenes.
\newblock \emph{ACM Trans. Graph.}, 36\penalty0 (4):\penalty0 130--1, 2017.

\bibitem[Li et~al.(2024)Li, Zhang, Zhang, Zhang, Li, Li, Ma, and Li]{li2024llavanext-interleave}
Feng Li, Renrui Zhang, Hao Zhang, Yuanhan Zhang, Bo Li, Wei Li, Zejun Ma, and Chunyuan Li.
\newblock Llava-next: Tackling multi-image, video, and 3d in large multimodal models, 2024.

\bibitem[Li et~al.(2023)Li, Li, Savarese, and Hoi]{li2023blip}
Junnan Li, Dongxu Li, Silvio Savarese, and Steven Hoi.
\newblock Blip-2: Bootstrapping language-image pre-training with frozen image encoders and large language models.
\newblock In \emph{International conference on machine learning}, pages 19730--19742. PMLR, 2023.

\bibitem[Lin et~al.(2023)Lin, Zhu, Ye, Ning, Jin, and Yuan]{lin2023video}
Bin Lin, Bin Zhu, Yang Ye, Munan Ning, Peng Jin, and Li Yuan.
\newblock Video-llava: Learning united visual representation by alignment before projection.
\newblock \emph{arXiv preprint arXiv:2311.10122}, 2023.

\bibitem[Lin(2004)]{lin2004rouge}
Chin-Yew Lin.
\newblock Rouge: A package for automatic evaluation of summaries.
\newblock In \emph{Text summarization branches out}, pages 74--81, 2004.

\bibitem[Liu et~al.(2024{\natexlab{a}})Liu, Li, Li, Li, Zhang, Shen, and Lee]{liu2024llavanext}
Haotian Liu, Chunyuan Li, Yuheng Li, Bo Li, Yuanhan Zhang, Sheng Shen, and Yong~Jae Lee.
\newblock Llava-next: Improved reasoning, ocr, and world knowledge, 2024{\natexlab{a}}.

\bibitem[Liu et~al.(2024{\natexlab{b}})Liu, Li, Wu, and Lee]{liu2024visual}
Haotian Liu, Chunyuan Li, Qingyang Wu, and Yong~Jae Lee.
\newblock Visual instruction tuning.
\newblock \emph{Advances in neural information processing systems}, 36, 2024{\natexlab{b}}.

\bibitem[Loshchilov(2017)]{loshchilov2017decoupled}
I Loshchilov.
\newblock Decoupled weight decay regularization.
\newblock \emph{arXiv preprint arXiv:1711.05101}, 2017.

\bibitem[Loshchilov and Hutter(2016)]{loshchilov2016sgdr}
Ilya Loshchilov and Frank Hutter.
\newblock Sgdr: Stochastic gradient descent with warm restarts.
\newblock \emph{arXiv preprint arXiv:1608.03983}, 2016.

\bibitem[Luo et~al.(2023)Luo, Zhao, Yang, Dong, Li, Lu, Wang, Hu, Qiu, and Wei]{luo2023valley}
Ruipu Luo, Ziwang Zhao, Min Yang, Junwei Dong, Da Li, Pengcheng Lu, Tao Wang, Linmei Hu, Minghui Qiu, and Zhongyu Wei.
\newblock Valley: Video assistant with large language model enhanced ability.
\newblock \emph{arXiv preprint arXiv:2306.07207}, 2023.

\bibitem[Maaz et~al.(2023)Maaz, Rasheed, Khan, and Khan]{maaz2023video}
Muhammad Maaz, Hanoona Rasheed, Salman Khan, and Fahad~Shahbaz Khan.
\newblock Video-chatgpt: Towards detailed video understanding via large vision and language models.
\newblock \emph{arXiv preprint arXiv:2306.05424}, 2023.

\bibitem[Maaz et~al.(2024)Maaz, Rasheed, Khan, and Khan]{maaz2024videogpt+}
Muhammad Maaz, Hanoona Rasheed, Salman Khan, and Fahad Khan.
\newblock Videogpt+: Integrating image and video encoders for enhanced video understanding.
\newblock \emph{arXiv preprint arXiv:2406.09418}, 2024.

\bibitem[Papineni et~al.(2002)Papineni, Roukos, Ward, and Zhu]{papineni2002bleu}
Kishore Papineni, Salim Roukos, Todd Ward, and Wei-Jing Zhu.
\newblock Bleu: a method for automatic evaluation of machine translation.
\newblock In \emph{Proceedings of the 40th annual meeting of the Association for Computational Linguistics}, pages 311--318, 2002.

\bibitem[Radford et~al.(2021)Radford, Kim, Hallacy, Ramesh, Goh, Agarwal, Sastry, Askell, Mishkin, Clark, et~al.]{radford2021learning}
Alec Radford, Jong~Wook Kim, Chris Hallacy, Aditya Ramesh, Gabriel Goh, Sandhini Agarwal, Girish Sastry, Amanda Askell, Pamela Mishkin, Jack Clark, et~al.
\newblock Learning transferable visual models from natural language supervision.
\newblock In \emph{International conference on machine learning}, pages 8748--8763. PMLR, 2021.

\bibitem[Shi et~al.(2022)Shi, Yang, Xu, Yuan, Li, Hu, and Zha]{shi2022emscore}
Yaya Shi, Xu Yang, Haiyang Xu, Chunfeng Yuan, Bing Li, Weiming Hu, and Zheng-Jun Zha.
\newblock Emscore: Evaluating video captioning via coarse-grained and fine-grained embedding matching.
\newblock In \emph{Proceedings of the IEEE/CVF conference on computer vision and pattern recognition}, pages 17929--17938, 2022.

\bibitem[Song et~al.(2024)Song, Chai, Wang, Zhang, Zhou, Wu, Chi, Guo, Ye, Zhang, et~al.]{song2024moviechat}
Enxin Song, Wenhao Chai, Guanhong Wang, Yucheng Zhang, Haoyang Zhou, Feiyang Wu, Haozhe Chi, Xun Guo, Tian Ye, Yanting Zhang, et~al.
\newblock Moviechat: From dense token to sparse memory for long video understanding.
\newblock In \emph{Proceedings of the IEEE/CVF Conference on Computer Vision and Pattern Recognition}, pages 18221--18232, 2024.

\bibitem[Soucek and Lokoc(2024)]{soucek2024transnet}
Tom{\'a}s Soucek and Jakub Lokoc.
\newblock Transnet v2: An effective deep network architecture for fast shot transition detection.
\newblock In \emph{Proceedings of the 32nd ACM International Conference on Multimedia}, pages 11218--11221, 2024.

\bibitem[Sun et~al.(2023)Sun, Fang, Wu, Wang, and Cao]{sun2023eva}
Quan Sun, Yuxin Fang, Ledell Wu, Xinlong Wang, and Yue Cao.
\newblock Eva-clip: Improved training techniques for clip at scale.
\newblock \emph{arXiv preprint arXiv:2303.15389}, 2023.

\bibitem[Team(2023)]{team2023internlm}
InternLM Team.
\newblock Internlm: A multilingual language model with progressively enhanced capabilities, 2023.

\bibitem[Truong et~al.(2016)Truong, Berthouzoz, Li, and Agrawala]{truong2016quickcut}
Anh Truong, Floraine Berthouzoz, Wilmot Li, and Maneesh Agrawala.
\newblock Quickcut: An interactive tool for editing narrated video.
\newblock In \emph{Proceedings of the 29th Annual Symposium on User Interface Software and Technology}, pages 497--507, 2016.

\bibitem[Vedantam et~al.(2015)Vedantam, Lawrence~Zitnick, and Parikh]{vedantam2015cider}
Ramakrishna Vedantam, C Lawrence~Zitnick, and Devi Parikh.
\newblock Cider: Consensus-based image description evaluation.
\newblock In \emph{Proceedings of the IEEE conference on computer vision and pattern recognition}, pages 4566--4575, 2015.

\bibitem[Wang et~al.(2019)Wang, Yang, Hu, Yau, Shamir, et~al.]{wang2019write}
Miao Wang, Guo-Wei Yang, Shi-Min Hu, Shing-Tung Yau, Ariel Shamir, et~al.
\newblock Write-a-video: computational video montage from themed text.
\newblock \emph{ACM Trans. Graph.}, 38\penalty0 (6):\penalty0 177--1, 2019.

\bibitem[Wang et~al.(2024)Wang, Li, Li, Yu, He, Chen, Pei, Zheng, Xu, Wang, et~al.]{wang2024internvideo2}
Yi Wang, Kunchang Li, Xinhao Li, Jiashuo Yu, Yinan He, Guo Chen, Baoqi Pei, Rongkun Zheng, Jilan Xu, Zun Wang, et~al.
\newblock Internvideo2: Scaling video foundation models for multimodal video understanding.
\newblock \emph{arXiv preprint arXiv:2403.15377}, 2024.

\bibitem[Xiong et~al.(2022)Xiong, Heilbron, and Lin]{xiong2022transcript}
Yu Xiong, Fabian~Caba Heilbron, and Dahua Lin.
\newblock Transcript to video: Efficient clip sequencing from texts.
\newblock In \emph{Proceedings of the 30th ACM International Conference on Multimedia}, pages 5407--5416, 2022.

\bibitem[Yang et~al.(2024{\natexlab{a}})Yang, Yang, Hui, Zheng, Yu, Zhou, Li, Li, Liu, Huang, et~al.]{yang2024qwen2}
An Yang, Baosong Yang, Binyuan Hui, Bo Zheng, Bowen Yu, Chang Zhou, Chengpeng Li, Chengyuan Li, Dayiheng Liu, Fei Huang, et~al.
\newblock Qwen2 technical report.
\newblock \emph{arXiv preprint arXiv:2407.10671}, 2024{\natexlab{a}}.

\bibitem[Yang et~al.(2024{\natexlab{b}})Yang, Zhan, Wang, Wang, Ge, Zheng, and Jin]{yang2024synchronized}
Dingyi Yang, Chunru Zhan, Ziheng Wang, Biao Wang, Tiezheng Ge, Bo Zheng, and Qin Jin.
\newblock Synchronized video storytelling: Generating video narrations with structured storyline.
\newblock \emph{arXiv preprint arXiv:2405.14040}, 2024{\natexlab{b}}.

\bibitem[Yao et~al.(2024)Yao, Yu, Zhang, Wang, Cui, Zhu, Cai, Li, Zhao, He, et~al.]{yao2024minicpm}
Yuan Yao, Tianyu Yu, Ao Zhang, Chongyi Wang, Junbo Cui, Hongji Zhu, Tianchi Cai, Haoyu Li, Weilin Zhao, Zhihui He, et~al.
\newblock Minicpm-v: A gpt-4v level mllm on your phone.
\newblock \emph{arXiv preprint arXiv:2408.01800}, 2024.

\bibitem[Zhai et~al.(2023)Zhai, Mustafa, Kolesnikov, and Beyer]{zhai2023sigmoid}
Xiaohua Zhai, Basil Mustafa, Alexander Kolesnikov, and Lucas Beyer.
\newblock Sigmoid loss for language image pre-training.
\newblock In \emph{Proceedings of the IEEE/CVF International Conference on Computer Vision}, pages 11975--11986, 2023.

\bibitem[Zhang et~al.(2023{\natexlab{a}})Zhang, Li, and Bing]{zhang2023video}
Hang Zhang, Xin Li, and Lidong Bing.
\newblock Video-llama: An instruction-tuned audio-visual language model for video understanding.
\newblock \emph{arXiv preprint arXiv:2306.02858}, 2023{\natexlab{a}}.

\bibitem[Zhang et~al.(2023{\natexlab{b}})Zhang, Wang, Cao, Xu, Ouyang, Zhao, Ding, Zhang, Duan, Yan, et~al.]{zhang2023internlm}
Pan Zhang, Xiaoyi Dong~Bin Wang, Yuhang Cao, Chao Xu, Linke Ouyang, Zhiyuan Zhao, Shuangrui Ding, Songyang Zhang, Haodong Duan, Hang Yan, et~al.
\newblock Internlm-xcomposer: A vision-language large model for advanced text-image comprehension and composition.
\newblock \emph{arXiv preprint arXiv:2309.15112}, 2023{\natexlab{b}}.

\bibitem[Zhang et~al.(2024)Zhang, Li, Liu, Lee, Gui, Fu, Feng, Liu, and Li]{zhang2024llavanextvideo}
Yuanhan Zhang, Bo Li, haotian Liu, Yong~jae Lee, Liangke Gui, Di Fu, Jiashi Feng, Ziwei Liu, and Chunyuan Li.
\newblock Llava-next: A strong zero-shot video understanding model, 2024.

\bibitem[Zhu et~al.(2023)Zhu, Chen, Shen, Li, and Elhoseiny]{zhu2023minigpt}
Deyao Zhu, Jun Chen, Xiaoqian Shen, Xiang Li, and Mohamed Elhoseiny.
\newblock Minigpt-4: Enhancing vision-language understanding with advanced large language models.
\newblock \emph{arXiv preprint arXiv:2304.10592}, 2023.

\end{thebibliography}
